\documentclass{article} 
\usepackage{iclr2019_conference,times}

\usepackage{hyperref}
\usepackage{url}

\title{Visual Explanation by Interpretation:\\Improving Visual Feedback Capabilities of Deep Neural Networks}

\author{
Jos{\'e} Oramas M.\footnotemark[1] \hspace{3cm} 
Kaili Wang\thanks{denotes equal contribution} \hspace{3cm}
Tinne Tuytelaars
\\
\hspace{5cm}\textit{KU Leuven, ESAT-PSI} 
}

\usepackage[utf8]{inputenc} 
\usepackage[T1]{fontenc}    
\usepackage{hyperref}       
\usepackage{url}            
\usepackage{booktabs}       
\usepackage{amsfonts}       
\usepackage{microtype}      
\usepackage{enumitem}
\usepackage{times}
\usepackage{epsfig}
\usepackage{graphicx}
\usepackage{amsmath}
\usepackage{amssymb}
\usepackage{subfig}
\usepackage{multicol}
\usepackage[nodisplayskipstretch]{setspace}
\usepackage{color} 
\newcommand{\aniFlower}[0]{{\small an8Flower}}
\newcommand{\MNIST}[0]{{\small MNIST}}
\newcommand{\ILSVRC}[0]{{\small ILSVRC}}
\newcommand{\Fashion}[0]{{\small Fashion}}
\newcommand{\VGG}[0]{{\small VGG}}
\newcommand{\F}[0]{{\small F~}}
\newcommand{\DNNs}[0]{{\small DNNs}}
\newcommand{\DNN}[0]{{\small DNN}}
\newcommand{\imageNet}[0]{{\small imageNet}}

\usepackage{xcolor}
\definecolor{OliveGreen}{RGB}{0,180,0}

\newcommand{\jose}[1]{\textcolor{black}{#1}}

\newcommand{\tinne}[1]{\textcolor{black}{#1}}

\iclrfinalcopy 
\begin{document}

\maketitle

\begin{abstract}
Interpretation and explanation of deep models is critical
towards \tinne{wide} adoption of systems that rely on them.
In this paper, we propose a novel scheme for both interpretation as well as explanation in which, given a 
pretrained model, we automatically identify internal features relevant 
for the set of classes
considered by the model, without 
\tinne{relying on} additional
annotations. We \textit{interpret} the model through average 
visualizations of this reduced set of features.
Then, at test time, we \textit{explain} the network prediction by 
accompanying the predicted class label with supporting 
visualizations derived from the identified features.
In addition, we propose a method to address the artifacts 
introduced by strided operations in deconvNet-based visualizations.
Moreover, we introduce \aniFlower, a dataset specifically designed 
for objective quantitative evaluation of methods for visual explanation.
Experiments on the \MNIST, \ILSVRC12, \Fashion144k and \aniFlower ~datasets 
show that our method produces detailed explanations with good coverage of 
relevant features of the classes of interest.

\end{abstract}



\section{Introduction}

Methods 
based on deep neural networks (\DNNs) have achieved impressive results for several 
computer vision tasks, such as
image classification,
object detection
and image generation.
Combined with the general tendency in the Computer Vision community of developing 
methods with a focus on high quantitative performance, 
this has motivated the 
\tinne{wide}
adoption of \DNN-based methods, 
despite 
\tinne{the initial skepticism due to}
their black-box characteristics. 
In this work, we aim for more visually-descriptive predictions and 
propose means to improve the quality of the visual feedback capabilities of 
\DNN-based methods. 
Our goal is to bridge the gap between methods aiming at model \textit{interpretation}, 
i.e.,~understanding what a given trained model has actually learned, and methods 
aiming at model \textit{explanation}, i.e.,~justifying the decisions made by a model.

\begin{figure}
  \centering
\includegraphics[width=0.98\textwidth]{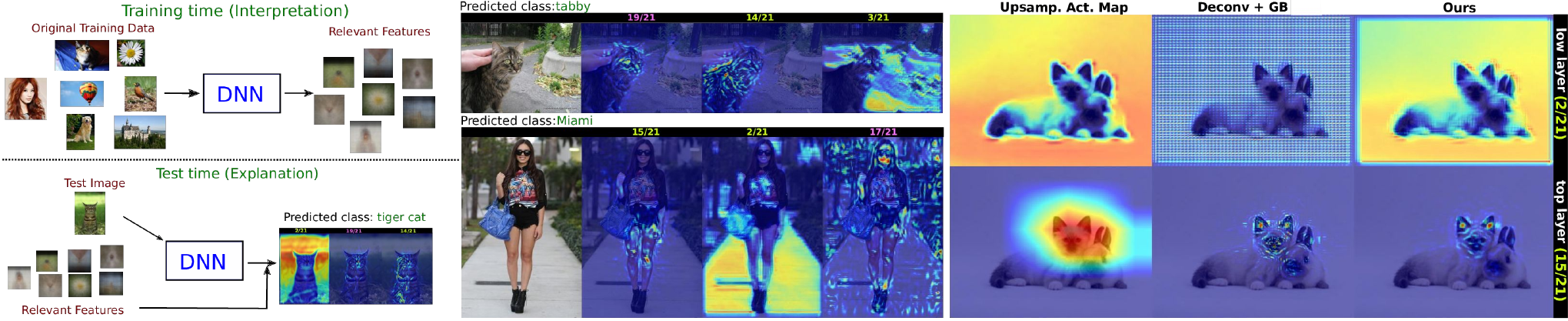}\\
\vspace{-2mm}
  \caption{ \small Left: Proposed training/testing pipeline. 
  Center: Visual explanations generated by our method.
  Predicted class labels are enriched with heatmaps indicating 
  the pixel locations, associated to the features, that contributed to 
  the prediction.
  Note these features may come from the object itself as well as from 
  the context. 
  On top of each heatmap we indicate the number of the layer where the features come
  from. The layer type is color-coded (green for convolutional and pink for fully connected).
  Right: Visualization comparison. Note how our heatmaps attenuate the grid-like artifacts introduced by  deconvnet-based methods at lower layers. At the same time, our method
is able to produce a more detailed visual feedback than up-scaled activation maps.
  }
   \label{fig:teaser}
  \vspace{-8mm}
\end{figure}

Model interpretation of \DNNs ~is commonly achieved in two ways: either by 
a) manually inspecting visualizations of every single filter~(or a random subset thereof) from every layer of the network~(\cite{YosinskiDeepVis15,ZeilerDeconv14})
or, more recently, by b) exhaustively comparing the internal activations produced by a 
given model w.r.t.~a dataset with pixel-wise annotations of possibly relevant concepts~(\cite{netdissect2017,fong18net2vec}).
These two paths have provided useful insights into the internal representations learned by \DNNs. 
However, they both have their own weaknesses.
For the first case, the manual inspection of filter responses introduces a subjective bias, 
as was evidenced by \cite{GonzalezGarcia2017DoSP}. In addition, the inspection of every 
filter from every layer becomes a cognitive-expensive practice for deeper models, 
which makes it a noisy process.
For the second case, as stated by \cite{netdissect2017}, the interpretation capabilities 
over the network are limited by the concepts for which annotation is available. Moreover, 
the cost of adding annotations for new concepts is quite high due to its pixel-wise nature. 
A third weakness, shared by both cases, is inherited by the way in which they
generate spatial filter-wise responses, i.e., either through deconvolution-based 
heatmaps~(\cite{SpringenbergGuidedBack15,ZeilerDeconv14}) or by up-scaling the 
activation maps at a given layer/filter to the image space~(\cite{netdissect2017,zhou2015cnnlocalization}).
On the one hand, 
deconvolution methods are able to produce heatmaps with 
high level of detail from any filter in the network. However, as can be seen in 
Fig.~\ref{fig:teaser}\,(right), they suffer from artifacts introduced by strided operations in the back-propagation process. 
Up-scaled activation maps, on the other hand, can significantly lose details
when displaying the response of filters with large receptive field from deeper layers. 
Moreover, they have the weakness of only being computable for convolutional layers.

In order to alleviate these issues, we start from the hypothesis proven by \cite{netdissect2017,YosinskiDeepVis15}, that \tinne{only} a small subset of the internal filters 
of a network encode features that are important for the task that the network addresses.
Based on that assumption, we propose a method which, given a trained 
\DNN ~model, automatically identifies a set of relevant internal filters whose encoded 
features serve as indicators for the class of interest to be predicted~(Fig.~\ref{fig:teaser}\, left). 
These filters can originate from any type of internal layer of the network, i.e.,~\textit{convolutional, 
fully connected}, etc. \tinne{Selecting them} is formulated as a \textit{$\mu$-lasso} optimization problem in 
which a sparse set of filter-wise responses are linearly combined in order to predict the 
class of interest. At test time, 
\tinne{we move from interpretation to explanation}.
Given an image, a set of identified relevant filters, 
and a class prediction, we accompany the predicted class label with heatmap visualizations
of the top-responding relevant filters for the predicted class, see Fig.~\ref{fig:teaser}\,(center).
In addition, by improving the resampling operations within deconvnet-based methods, 
we are able to address the artifacts introduced in the back-propagation process, see Fig.~\ref{fig:teaser}\,(right).
The code and models used to generate our visual explanations can be found in the following link~\footnote{\scriptsize{\url{http://homes.esat.kuleuven.be/~joramas/projects/visualExplanationByInterpretation}}}.
Overall, the proposed method removes the requirement of additional expensive 
pixel-wise annotation, by relying on the same annotations used to train the initial model.
Moreover, by using our own variant of a deconvolution-based method, our method is able to 
consider the spatial response from any filter at any layer while still providing visually 
pleasant feedback. This allows our method to reach some level of explanation by interpretation.

Finally, recent approaches to evaluate explanation methods 
measure the validity of an explanation either via user studies~(\cite{ZeilerDeconv14,SelvarajuDVCPB16}) 
or by measuring its effect on a proxy task, e.g. object detection/segmentation~(\cite{Zhou15Parts,zhang2016EB}).
While user studies inherently add subjectivity, 
benchmarking through a proxy task steers the optimization 
of the explanation method towards such task.
Here we propose an objective evaluation via \textit{\aniFlower}, a synthetic 
dataset where the discriminative feature between the classes of interest 
is controlled. 
This allows us to produce ground-truth masks for the regions 
to be highlighted by the explanation. Furthermore, it allows us to 
quantitatively measure the performance of methods for model explanation.

The main contributions of this work are four-fold.
First, we propose an automatic method based on feature selection 
       to identify the network-encoded features that are important for the 
       prediction of a given class. This alleviates the requirement of exhaustive 
       manual inspection or additional expensive pixel-wise annotations required by 
       existing methods. 
Second, the proposed method is able to provide visual feedback with higher-level 
       of detail over up-scaled raw activation maps 
       and improved quality over recent deconvolution+guided 
       back-propagation methods. 
Third, the proposed method is general enough to be applied to any type of network,
       independently of the type of layers that compose it.
Fourth, we release a dataset and protocol specifically 
        designed for the evaluation of methods for model explanation. 
       To the best of our knowledge this is the first dataset aimed at such task.

This paper is organized as follows: in Sec.~\ref{sec:relatedWork} 
we position our work w.r.t. existing work. Sec.~\ref{sec:proposedMethod} 
presents the pipeline and inner-workings of the proposed method.
In Sec.~\ref{sec:evaluation}, we conduct a series of experiments 
evaluating different aspects of the proposed method.
We draw conclusions in Sec.~\ref{sec:conclusion}.


\section{Related Work}
\label{sec:relatedWork}

\textbf{Interpretation.}
\cite{ZeilerDeconv14,Zhou15Parts} proposed to visualize properties 
of the function modelled by a network by systematically covering
(part of) the input image and measuring the difference of activations.  
The assumption is that occlusion of important parts of the 
input will lead to a significant drop in performance.
This procedure is applied at test time to identify the 
regions of the image that are important for classification.
However, the resolution of the explanation will depend on the region 
size.
%
Another group of works focuses on linking internal activations 
with semantic concepts. \cite{escorciaCVPR15} proposed a feature selection 
method in which the neuron activations of a \DNN ~trained with object 
categories are combined to predict object attributes.
Similarly, \cite{netdissect2017,fong18net2vec,ZhangInterCNN17} 
proposed to exhaustively 
match the activations of every filter from the convolutional layers
against a dataset with pixel-wise annotated concepts.
While both methods provide important insights on the semantic concepts 
encoded by the network, they are both limited by the concepts 
for which annotation is available. 
Similar to \cite{escorciaCVPR15}, we discover relevant internal 
filters through a feature selection method. Different from it,
we link internal activations directly to the same annotations 
used to train the initial model. This removes the expensive 
requirement of additional annotations.
%
A third line of works aims at discovering frequent visual 
patterns~(\cite{doersch2015makes, RematasCVPR15}) 
occurring in image collections.
These patterns have a high semantic coverage which makes 
them effective as means for summarization.
We  adopt the idea of using visualizations of (internal) mid-level 
elements as means to reveal the relevant features encoded, internally, 
by a \DNN. More precisely, we use the average visualizations used by these
works in order to interpret, visually, what the network has actually learned.

\textbf{Explanation.}
For the sake of brevity, we ignore methods which generate explanations via bounding boxes~(\cite{KarpathyVisualAligments15,oramasVisualComp16}) or 
text~(\cite{HendricksImageDescr16}), and focus on methods capable of generating 
visualizations with pixel-level precision.
\cite{ZeilerDeconv14} proposed a deconvolutional network~(Deconvnet) which uses activations from a given top layer and reverses the forward pass to reveal which visual patterns from the input image are responsible for the observed activations. \cite{Simonyan14} used information from the lower layers and the input image to estimate which image regions are responsible for the activations seen at the top layers.
Similarly, \cite{bachLRP15} decomposed the classification decision into 
pixel-wise contributions while preserving the propagated quantities 
between adjacent layers.
Later, \cite{SpringenbergGuidedBack15} extended these works by 
introducing ``guided back-propagation'', a technique that removes the effect of units with negative contributions 
in forward and backward pass. This resulted in sharper heatmap visualizations. 
\cite{zhou2015cnnlocalization} propose Global Average Pooling, i.e., 
a weighted sum over the spatial locations of the activations of the filters of the last convolutional layer, which results in a class activation map. Finally, a heatmap is generated by upsampling the class activation map to the size of the input image.
\cite{SelvarajuDVCPB16} extended this by providing a more efficient way for computing the weights for the activation maps. Recently, \cite{ChattopadhyayGCPP} extended this with neuron specific weights with the goal of improving object localization on the generated visualizations.
Here, we take DeconvNet with guided-backpropagation as starting point given its
maturity and ability to produce visual feedback with pixel-level precision. 
However, we change the internal operations in the backward pass with the goal of reducing 
visual artifacts introduced by strided operations while maintaining the network structure.

\textbf{Benchmarking.}
%
\cite{zhou2015cnnlocalization,zhang2016EB} proposed a saliency-based evaluation 
where explanations are assessed based on how well they highlight complete instances of the classes of 
interest. Thus, treating model explanation as a weakly-supervised object detection/segmentation problem. 
This saliency-based protocol assumes that explanations are exclusive to intrinsic object features, 
e.g. color, shape, parts, etc. and completely ignores extrinsic features, e.g. scene, environment, 
related to the depicted context. 
\cite{ZeilerDeconv14,SelvarajuDVCPB16} proposed a protocol 
based on crowd-sourced user studies. These type of evaluations are not only characterized 
by their high-cost, but also suffer from subjective bias (\cite{GonzalezGarcia2017DoSP}).
Moreover, \cite{vqahat} suggest that deep models and humans do not necessarily attend to 
the same input evidence even when they predict the same output.
Here we propose a protocol where the regions to be highlighted by the explanation 
are predefined. The goal is to \textit{objectify} the evaluation and relax 
the subjectivity introduced by human-based evaluations. Moreover, our protocol 
makes no strong assumption regarding the type of features highlighted by the
the explanations.

\section{Proposed Method}
\label{sec:proposedMethod}

The proposed method consists of two parts. At training time, a set of relevant 
layer/filter pairs are identified for every class of interest $j$. 
This results in a relevance weight $w_j$, associated to class $j$, for every 
filter-wise response $x$ computed internally by the network. 
At test time, an image $I$ is pushed through the network producing the class 
prediction {\small $\hat{j} {=} F(I)$}. Then, taking into account the internal responses 
$x$, and relevance weights $w_{\hat{j}}$ for the predicted class $\hat{j}$, we generate visualizations indicating the image regions that contributed to this prediction. 


\subsection{Identifying Relevant Features}
\label{sec:identifyRelevanFeatures}

One of the strengths of deep models is
their ability to learn abstract concepts from simpler ones. 
That is, when an example is pushed into the model, a conclusion 
concerning a specific task can be reached as a function of the 
results (activations) of intermediate operations at different 
levels (layers) of the model.
These intermediate results may hint at the ``semantic'' concepts 
that the model is taking into account when making a decision.
From this observation, 
and the fact that activations are typically sparse,
we make the assumption that some 
of the internal filters of a network encode features that are 
important for the task that the network addresses.
To this end, we follow a procedure similar to \cite{escorciaCVPR15}, 
aiming to predict each class $j$ by the linear combination {\small$w_j{\in} \mathbb{R}^m$} of its internal activations $x$, with $m$ the total number of neurons/activations.

As an initial step, we extract the image-wise response $x_i$. 
To this end, we compute the $L_2$ norm of each channel (filter response) within 
each layer and produce a 1-dimensional descriptor by concatenating 
the responses from the different channels. This layer-specific descriptor is 
$L_1$-normalized in order to compensate for the difference in 
length among different layers. Finally, 
we concatenate all the layer-specific descriptors
to obtain $x_i$.
%
In this process, we do not consider the last layer whose output is directly 
related to the classes of interest, e.g. the last two layers from \VGG-\F~\cite{Chatfield14}.

Following this procedure, we construct the matrix {\small $X {\in} \mathbb{R}^{m\times N}$} 
by passing each of the {\small $N$} training images through 
the network {\small $F$} and storing the internal responses $x$. 
As such, the {\small $i^{th}$} image of the dataset is 
represented by a vector {\small $x_i {\in} \mathbb{R}^m$} defined by the 
filter-wise responses at different layers. 
Furthermore, the possible classes that the {\small $i^{th}$} image 
belongs to are organized in a binary vector {\small $l_i {\in} \{0,1\}^C$} 
where $C$ is the total number of classes. 
Putting the annotations from all the images together produces the binary 
label matrix {\small $L {=} [ l_1, l_2, ..., l_N ]$}, with {\small $L {\in} \mathbb{R}^{C\times N}$}.
With these terms, we resort to solving the equation:
\vspace{-2mm}
\begin{equation} 
\label{eq:matrixDecom}
\begin{aligned}
W^{*} &= argmin_W~||X^TW - L^T||^2_F 
 &~subject~to:~ ||w_j||_1 \leq \mu ~,~ \forall_j=1,...,C \\
\end{aligned}
\end{equation} 
%
%
with $\mu$ a parameter that allows controlling the sparsity. This is the matrix form of the $\mu$-lasso problem. 
This problem can be efficiently solved using the Spectral 
Gradient Projection method~\cite{SPAMSMairal,lassoSPG}.
The $\mu$-lasso formulation is optimal for cases like the ones obtained 
by ResNet where the number of internal activations is large compared to 
the number of examples.
After solving the $\mu$-lasso problem, we have a 
matrix {\small $W{=}[w_1,w_2,...,w_C]$}, with {\small $W {\in} \mathbb{R}^{m \times C}$}.
We impose sparsity on $W$ by enforcing the constraints on 
the $L_1$ norm of $w_j$, i.e., $||w_j||_1 {\leq} \mu ~,~ \forall_j=1,...,C $. 
As a result, each non-zero element in $W$ represents a 
pair of network layer $p$ and filter index $q$ (within the layer) 
of relevance.


\subsection{Generating Visual Feedback}

During training time (Section~\ref{sec:identifyRelevanFeatures}), 
we identified a set of relevant features (indicated by {\small $W$})
for the classes of interest. At test time, we generate the feedback 
visualizations by taking into account the response of these
features on the content of the tested images.
Towards this goal, we push an image {\small $I$} through the network producing the class 
prediction {\small $\hat{j} {=} F(I)$}. During that pass, we compute the internal 
filter-wise response vector $x_i$ following the procedure presented above.
Then we compute the response {\small $r^{\hat{j}}_i {=} (w_{\hat{j}} \circ x_i)$}, where $\circ$ represents the element-wise product between two vectors.
Note that the $w_{\hat{j}}$ vector is highly sparse, therefore adding an insignificant cost at test time.
The features, i.e., layer/filter pairs $(p^\ast,q^\ast)$, with strongest contribution in the 
prediction $\hat{j}$ are selected as those with maximum response in $r^{\hat{j}}_i$.
Finally, we feed this information to the Deconvnet-based method with 
guided backpropagation from \cite{gruen16} to visualize the important 
features as defined by the layer/filter pairs $(p^\ast,q^\ast)$. 
Following the visualization method from \cite{gruen16}, given a
filter $p$ from layer $q$ and an input image, we first push forward 
the input image through the network, storing the activations from 
each filter at each layer, until reaching the layer $p$.
Then, we backpropagate the activations from filter $q$ at layer $p$ 
with inverse operations until reaching back to the input image space. 
As result we get as part of the output a set of heatmaps, associated 
to the relevant features, defined by {\small $(p^\ast,q^\ast)$}, indicating 
the influence of the pixels that contributed to the prediction.
See Fig.\ref{fig:teaser}\,(left) for an example of the visual feedback 
provided by our method.
Please refer to \cite{gruen16,SpringenbergGuidedBack15,ZeilerDeconv14} 
for further details regarding Deconvnet-based and Guided backpropagation methods.


\begin{figure}
\begin{minipage}{0.60\linewidth}
  \includegraphics[width=\textwidth]{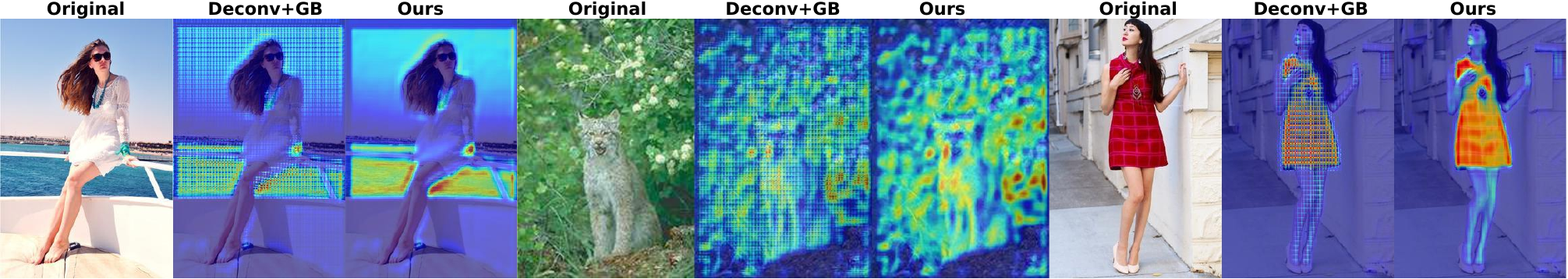}
  \vspace{-6mm}
  \caption{\small Heatmap visualization at lower layers of \VGG-\F.
   Note how our method attenuates the grid-like artifacts introduced 
   by existing DeconvNet+GB methods~(\cite{SpringenbergGuidedBack15}). 
   }
\label{fig:visualQuality}
\end{minipage}
\hfill
\begin{minipage}{0.37\linewidth}
\includegraphics[width=\textwidth]{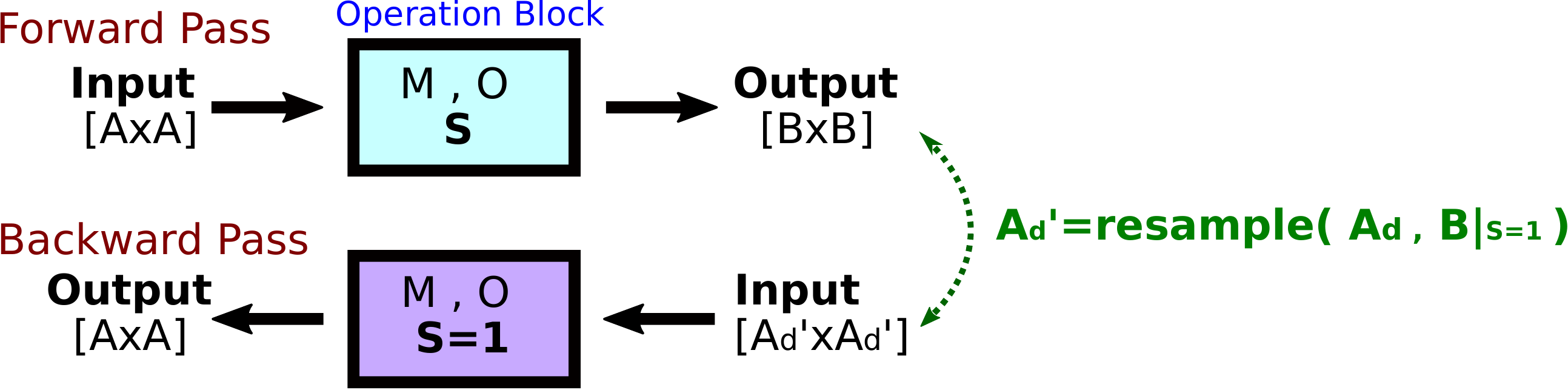}
\vspace{-6mm}
\caption{\small To attenuate artifacts 
, during the backward pass, we set the stride to 1 ($S=1$) and compensate 
by resampling the input so that $A_d'=B|_{S=1}$. 
}
\label{fig:ourDeconvMethod}
\end{minipage}%
\vspace{-6mm}
\end{figure}

\subsection{Improving Visual Feedback Quality}
\label{sec:improvingVisualFeedback}

Deep neural networks addressing computer vision tasks 
commonly push the input visual data through a sequence of operations.
A common trend of this sequential processing is that the input data is 
internally resampled until reaching the desired prediction space. 
As mentioned in Sec.~\ref{sec:relatedWork}, methods aiming at 
interpretation/explanation start from an internal point in the network 
and go backwards until reaching the input space - producing a heatmap.
However, due to the resampling process, heatmaps generated by the 
backwards process tend to display grid-like artifacts. 
More precisely, we find that this grid effect is caused by the internal 
resampling introduced by network operations with stride larger than one 
({\footnotesize  $S{>}1$}). 
To alleviate this effect, in the backwards pass, we set the stride 
{\footnotesize $S{=}1$} and compensate for this change by modifying the input accordingly.
As a result, the backwards process can be executed while maintaining 
the network structure.

More formally, given a network operation block defined by a convolution 
mask with size {\footnotesize $[M {\times} M]$}, stride {\footnotesize $[S,S]$}, and padding {\footnotesize $ [O,O,O,O]$}, 
the relationship between the size of its input {\footnotesize $[A {\times} A]$} and its output 
{\footnotesize $[B {\times} B]$} (see Fig.~\ref{fig:ourDeconvMethod}) is characterized 
by the following equation:
%
%
\begin{equation}
A + 2 \cdot O = M + (B - 1) \cdot S
\label{eq:inputImgSize}
\vspace{-2mm}
\end{equation}
from where, 
  \vspace{-2mm}
\begin{equation}
B = [~(A + 2 \cdot O - M)/S~] + 1
\label{eq:feaMapSize}
\end{equation}
Our method starts from the input~({\footnotesize $[A_d {\times} A_d]$}), which encodes the contributions from the input image, carried by the higher layer in the Deconvnet backward pass. 
In order to enforce a ``cleaner'' resampling when {\footnotesize $S{>}1$}, during the backward pass, the size of the input~({\footnotesize $[A_d {\times} A_d]$}) of the operation block should be the same as that of the feature map ({\footnotesize $[B {\times} B]$}) produced by the forward pass
if the stride {\footnotesize $S$} were equal to one, i.e., {\footnotesize $A'_d {=} B|_{S{=}1}$}. 
According to Eq.~\ref{eq:feaMapSize} with {\footnotesize $\footnotesize S{=}1$},  {\footnotesize $A_d$} 
should therefore be resampled to {\footnotesize $ A'_d {=} B|_{S{=}1} {=} A {+} 2 \cdot O  {-} M {+} 1 
$}. 
We do this resampling via  the nearest-neighbor interpolation algorithm given its 
proven fast computation time which makes it optimal for real-time processing.  
%
By introducing this step, the network will perform the backwards pass \jose{at every layer} 
with stride {\footnotesize $S{=}1$} and the grid effect will disappear. 
See Fig.~\ref{fig:visualQuality} for some examples of the 
improvements introduced by our method.


\section{Evaluation}
\label{sec:evaluation}

We conduct four sets of experiments.
First, in Sec.~\ref{sec:importanceRelFeat}, we verify the 
importance of the identified relevant features in the task addressed by the 
network.
Then, in Sec.~\ref{sec:visualFeedbackQuality}, we 
evaluate the improvements on visual quality provided by our method. 
In Sec.~\ref{sec:visualExplanationBenchmark}, 
we quantify the capability of our visual explanations to highlight the regions 
that are descriptive for the classes of interest.
Finally, in Sec.~\ref{sec:sanityCheck}, we assess the sensitivity 
of the proposed method w.r.t. the predicted classes.

We evaluate the proposed method on an image recognition task.
We conduct experiments on three standard image recognition 
datasets, i.e., \MNIST~\cite{lecunMNIST10}, \Fashion144k~\cite{SimoSerra15} 
and \imageNet~(\ILSVRC'12)~\cite{ILSVRC15}.
Additionally, we conduct experiments on a subset of cat images from 
\imageNet ~(\imageNet-cats).
\MNIST ~covers 10 classes of hand-written digits. It is 
composed by 70k images in total~(60k for training/validation, 10k for testing). 
The \imageNet ~dataset is composed of 1k classes. Following the standard practice, 
we measure performance on its validation set. Each class contains 50 validation 
images.
For the \Fashion144k ~dataset~\cite{SimoSerra15}, we consider the subset of 12k images  
from \cite{wangGeo17} used for the geolocation of 12 city classes. 
The \imageNet-cats subset consists of 13 cat classes, containing both 
domestic and wild cats. It is composed of 17,550 images. Each class contains 1,3k 
images for training and 50 images for testing.
Please refer to the supplementary material for more implementation details.

\begin{figure*}[!tbp]
\centering
    \includegraphics[width=0.8\textwidth]{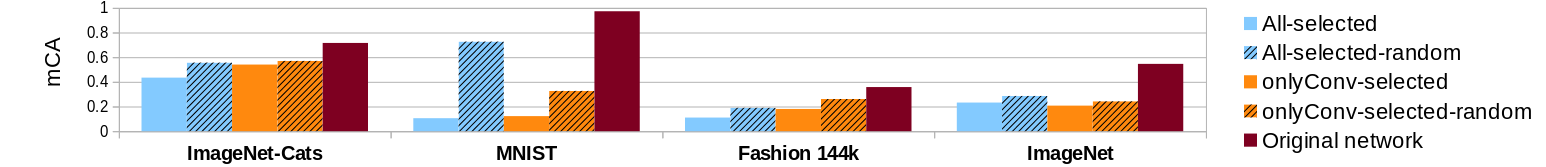}
    \vspace{-2mm}
  \caption{\small Changes in mean classification accuracy (mCA) as the identified relevant filters are ablated.}
  
  \label{fig:ablationExp}
  \vspace{-4mm}  
\end{figure*}

\subsection{Importance of Identified Relevant Features}
\label{sec:importanceRelFeat}


In this experiment, we verify the importance of the ``relevant'' features
identified by our method at training time~(Sec.~\ref{sec:identifyRelevanFeatures}).
To this end, given a set of identified features,  we evaluate the influence they 
have in the network by measuring changes in classification performance caused 
by their removal. We remove features in the network by setting their 
corresponding layer/filter to zero. The expected behavior is that a set of 
features with high relevance will produce a stronger drop in performance 
when ablated.
Fig.~\ref{fig:ablationExp}, shows the changes in classification 
performance for the tested datasets. We report the performance of four
sets of features: a)~\textit{All}, selected with our method by considering the whole
internal network architecture, b)~\textit{OnlyConv}, selected by considering 
only the convolutional layers of the network, c) a ~\textit{Random} 
selection of features~(filters) selected from the layers indicated in 
the sets a) and b), and for reference, d) the performance obtained 
by the original network.
Note that the \textit{OnlyConv} method makes the assumption 
that relevant features are only present in the convolutional layers. 
This is a similar assumption as the one made by state-of-the-art methods \cite{netdissect2017,XieNIPS17,zhou2015cnnlocalization}.
When performing feature selection (Sec.\ref{sec:identifyRelevanFeatures}), we set the sparsity parameter {\small$\mu{=}10$} for all the tested datasets. 
This produces subsets of {\small 92$\mid$101}, {\small 46$\mid$28}, {\small 104$\mid$111}, {\small 248$\mid$180} relevant features for the \textit{All$\mid$OnlyConv} methods, on the respective datasets
from Fig.~\ref{fig:ablationExp}.
Differences in the number of the selected features can be attributed 
to possibly redundant or missing predictive information between the initial pools 
of filter responses $x$ used to select the \textit{All} and \textit{OnlyConv}
features. 

A quick inspection of Fig.~\ref{fig:ablationExp} shows that indeed classification 
performance drops when we remove the identified features, \textit{All} and 
\textit{OnlyConv}. Moreover, it is noticeable that a random removal of 
features has a lower effect on classification accuracy.
This demonstrates the relevance of the identified features for the 
classes of interest.
In addition, it is visible that the method that considers the complete 
internal structure, i.e., \textit{All}, suffers a stronger drop in 
performance compared to the \textit{OnlyConv} which only considers features 
produced by the convolutional layers. This suggests that there is indeed 
important information encoded in the fully connected layers, and while 
convolutional layers are a good source for features, focusing on them only
does not reveal the full story.
\jose{
Regarding the effect of the sparsity value $\mu$ in the $\mu$-lasso 
formulation (Sec.~\ref{sec:identifyRelevanFeatures}), 
we note that increasing $\mu$  increases the number of selected features. This leads to more specialized 
features that can 
better cope with
rare instances of the classes 
of interest. We decided to start from a relatively low value, e.g. $\mu{=}10$, 
in order to focus on a small set of relevant features that can generalize to 
the classes of interest while, at the same time, keeping 
the computational cost low.
}

\begin{figure}
  \centering
  \includegraphics[width=1\textwidth]{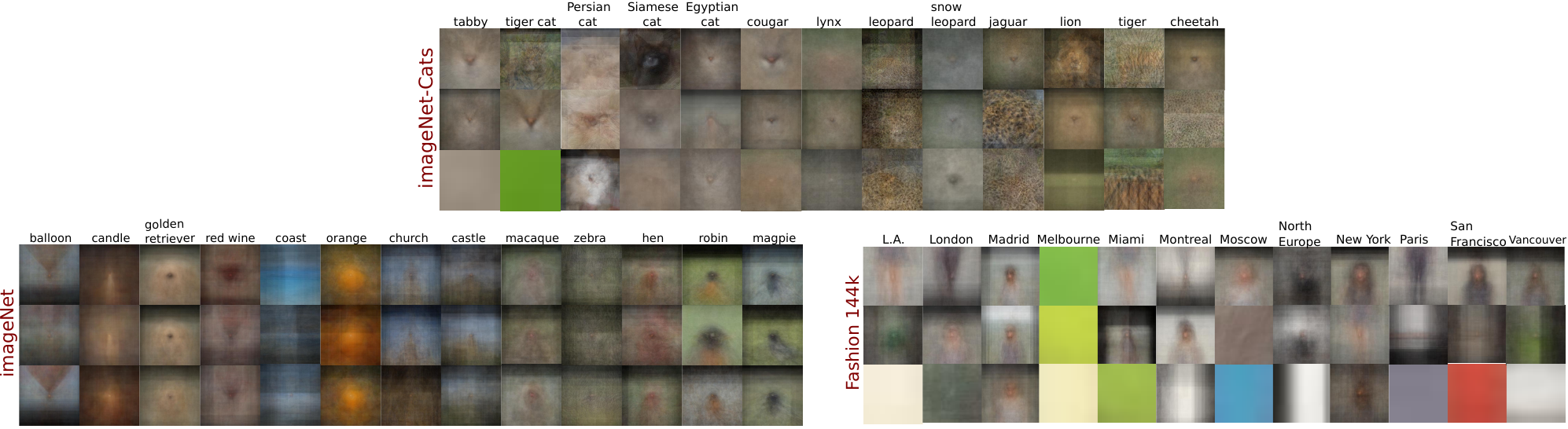}
  \vspace{-6mm}
  \caption{\small Average Images from the identified relevant filters for the ImageNet-Cats subset~(top),  some selected classes from the full ImageNet~(left) and the \Fashion144K~(left) datasets, respectively.}
  
  \label{fig:avgImages}
  \vspace{-6mm}
\end{figure}

\begin{figure}
  \centering
  \includegraphics[width=1\textwidth]{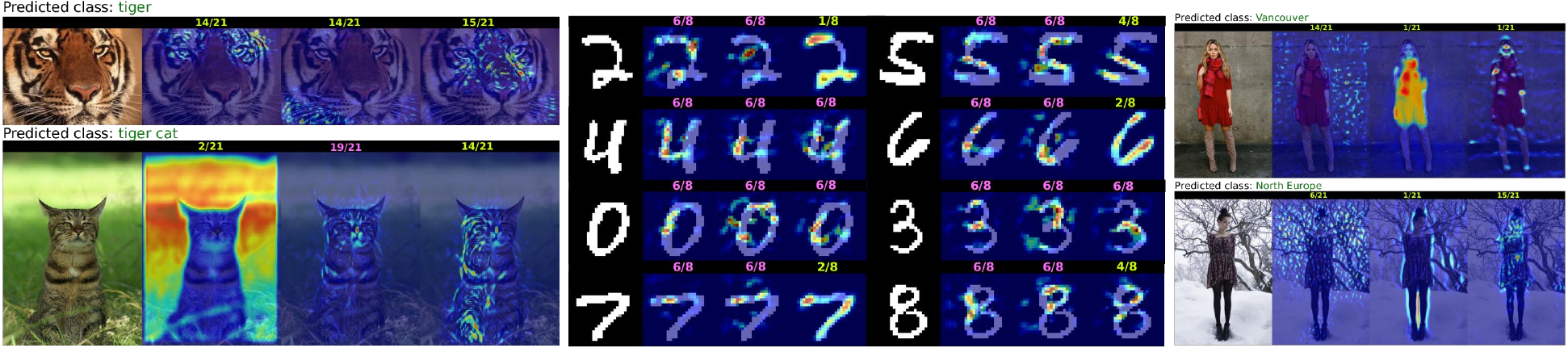}
   \vspace{-6mm}
  \caption{\small Our visual explanations.
  We accompany the predicted class label with our 
  heatmaps indicating the pixel locations, associated to the features, 
  that contributed to the prediction.
  These features may come from the object itself as well as 
  from its context. 
  See how for \MNIST, some features support the 
  existence of gaps, as to avoid confusion with another class.
  On top of each heatmap we indicate the number of the layer where the features come
  from. The layer type is color-coded, i.e., convolutional (green) and fully connected (pink).}
  
  \label{fig:visualExplanationImages}
  \vspace{-6mm}
\end{figure}

\textbf{Qualitative Analysis.}
In order to get a qualitative insight into the type of information 
that these features encode we compute an average visualization
by considering the top 100 image patches where such features have a high response.
Towards this goal, given the set of identified relevant features, for 
every class, we select images with higher responses. Then, we 
take the input image at the location with maximum response for a 
given filter and crop it by considering the receptive field of the corresponding 
layer/filter of interest.
Selected examples of average images, with rich semantic representation, 
are presented in Fig.~\ref{fig:avgImages} for the tested datasets.

We can notice that for imageNet-Cats, the identified features cover 
descriptive characteristics of the considered cat classes. For example, 
the dark head of a Siamese cat, the nose/mouth of a 
cougar, or the fluffy-white body shape of Persian cat.
Likewise, it effectively identifies the descriptive fur patterns from 
the jaguar, leopard and tiger classes and colors which are related to the background.
We see a similar effect on a selection of other objects from the rest of
the imageNet dataset.
For instance, for scene-type classes, i.e., coast, castle and church, the 
identified features focus on the outline of such scenes.
Similarly, we notice different viewpoints for animal-type classes, e.g.~golden-retriever, hen, robin, magpie.
On the \Fashion144k dataset (Fig.~\ref{fig:avgImages}\,(right)) 
we can notice that some classes respond to features related to 
green, blue, red, and beige colors. 
Some focus on legs, covered and uncovered, while others focus 
on the upped body part. It is interesting that from the upper 
body parts, some focus on persons with dark long hair, short hair, 
and light hair. Similarly, there is a class with high response to
horizontal black-white gradients where individuals tend to dress
in dark clothes.
These visualizations answer the question explored in 
\cite{wangGeo17} and why the computer outperforms the surveyed 
participants.  It shows that the model effectively exploits 
human-related features (legs clothing, hair length/color, clothing color) 
as well as background-related features, mainly covered by 
color/gradients and texture patterns.
In the visual explanations provided by our method 
we can see that the model effectively 
uses this type of features to reach its decision.

Finally, in Fig.~\ref{fig:visualExplanationImages} we show some examples
of the visual explanations produced by our method.
We aggregate the predicted class label with our heatmap visualizations 
indicating the pixel locations, associated to the relevant features, 
that contributed to the prediction.
For the case of the \ILSVRC'12 and \Fashion144k examples, we notice that the relevant 
features come from the object itself as well as from its context. 
For the case of the \MNIST~examples, in addition to the features firing 
on the object, there are features that support the existence of a gap 
(background), as to emphasize that the object is not filled there and 
avoid confusion with another class. 
For example, see for class $2$ how it speaks against $0$ and for $6$ how it goes against $4$.

\vspace{-1mm}
\subsection{Visual Feedback Quality}
\label{sec:visualFeedbackQuality}
\begin{figure}
\begin{minipage}{0.7\linewidth}

  \includegraphics[width=\textwidth]{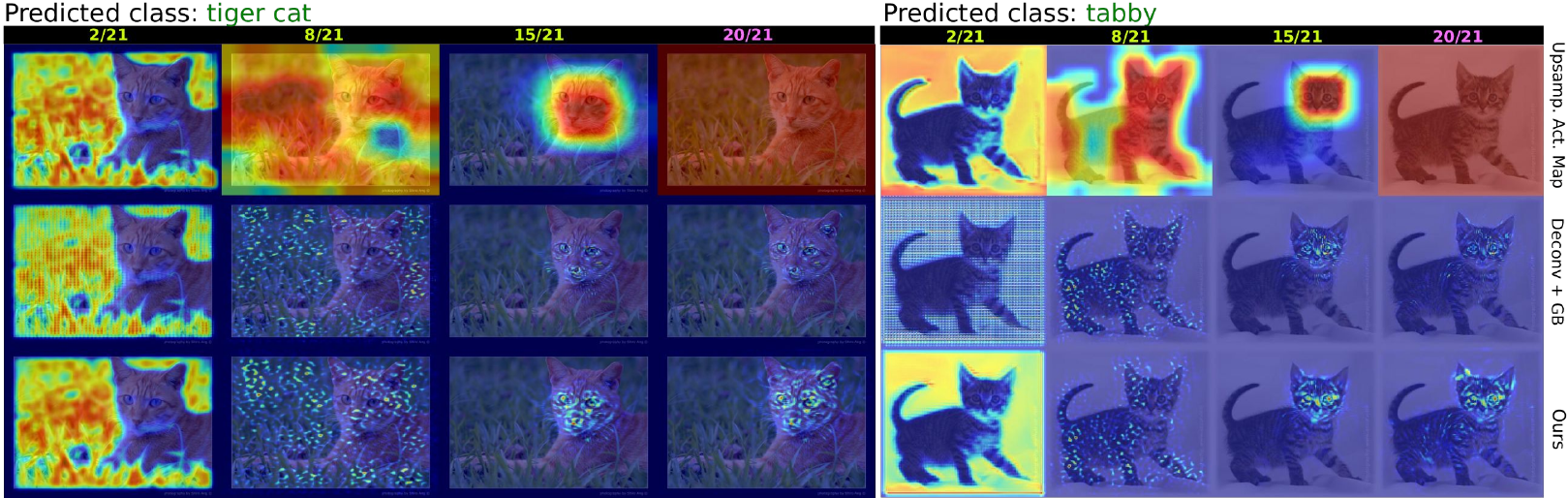}
   \vspace{-6mm}
  \caption{\small Pixel effect visualization for different methods. Note
   how for lower layers (8/21), our method attenuates the grid-like artifacts  
   introduced by Deconvnet methods. For higher layers (15/21), 
   our method provides a more precise visualization when compared to 
   upsampled activation maps. For the case of FC layers (20/21), using upsampled activation maps is not applicable.
   }
  \label{fig:visualQualityExp}

\end{minipage}
%
\hfill
%
\begin{minipage}{0.29\linewidth}

  \resizebox{0.98\columnwidth}{!}{%

  \begin{tabular}{lc}
  
    \toprule
    Method     			& \textbf{single-6c}  \\
    \midrule
    Upsam. Act.									& 16.8$\pm$2.63	 \\
    Deconv+GB,~\cite{SpringenbergGuidedBack15}	& 21.3$\pm$0.77	 \\
    Grad-CAM,~\cite{vqahat}						& 17.5$\pm$0.25  \\
    Guided Grad-CAM,~\cite{vqahat}		        & 19.9$\pm$0.61	 \\
    Grad-CAM++,~\cite{ChattopadhyayGCPP}			& 15.6$\pm$0.57  \\
    Guided Grad-CAM++,~\cite{ChattopadhyayGCPP}	& 19.6$\pm$0.65  \\
    \textbf{\textit{Ours}}						& \textbf{22.5$\pm$0.82}   \\
    
    \midrule
    
    Method     			& \textbf{double-12c} \\
    \midrule
    Upsam. Act.									& 16.1$\pm$1.30 \\
    Deconv+GB,~\cite{SpringenbergGuidedBack15}	& 21.9$\pm$0.72 \\
    Grad-CAM,~\cite{vqahat}						& 14.8$\pm$0.16 \\
    Guided Grad-CAM,~\cite{vqahat}		        & 19.4$\pm$0.34 \\
    Grad-CAM++,~\cite{ChattopadhyayGCPP}			& 14.6$\pm$0.12  \\
    Guided Grad-CAM++,~\cite{ChattopadhyayGCPP}	& 19.7$\pm$0.27  \\    
    \textbf{\textit{Ours}}						& \textbf{23.2$\pm$0.60} \\

  \end{tabular}
  \label{tab:synExpAUC}

  }

\caption*{\small Table 1: Area under the IoU curve (in percentages) on an8Flower over \textit{5-folds}.}

\end{minipage}%

\vspace{-5mm}

\end{figure}

\vspace{-2mm}
In this section, we assess the visual quality of the
visual explanations generated by our method. 
In Fig.~\ref{fig:visualQualityExp}, we compare 
our visualizations with upsampled activation maps from 
internal layers~(\cite{netdissect2017,zhou2015cnnlocalization}) 
and the output of DeconvNet with guided-backpropagation~(\cite{SpringenbergGuidedBack15}).
We show these visualizations for different layers/filters  
throughout the network.

A quick inspection reveals that our method to attenuate the grid-like 
artifacts introduced by Deconvnet methods (see Sec~\ref{sec:improvingVisualFeedback}) indeed produces noticeable improvements for lower layers. 
See Fig.~\ref{fig:visualQuality} for additional examples presenting this 
difference at lower layers.
Likewise, for the case of higher layers~(Fig.~\ref{fig:visualQualityExp}), 
the proposed method provides more precise visualizations when compared 
to upsampled activation maps. 
In fact, the rough output produced by the activation maps at higher 
layers has a saliency-like behavior that gives the impression that the 
network is focusing on a larger region of the image. This could be a 
possible attribution to why in earlier works \cite{Zhou15Parts}, 
manual inspection of network activations suggested that the network 
was focusing on ``semantic'' parts. 
Please see~\cite{GonzalezGarcia2017DoSP} for an in-depth discussion
of this observation.
Finally, for the case of FC layers, using upsampled activation maps 
is not applicable.
\noindent Please refer to the supplementary material for additional examples.
In addition, to quantitatively measure the quality of our heatmaps we perform 
a box-occlusion study~\cite{ZeilerDeconv14}. Given a specific heatmap, we 
occlude the original image with patches sampled from the distribution defined 
by the heatmap. We measure changes in performance as we gradually increase the 
number of patches up to covering the 30\% most relevant part of the image.
Here our method reaches a mean difference in prediction confidence of 2\% w.r.t. to 
\cite{SpringenbergGuidedBack15}. This suggests that our method is able to maintain 
focus on relevant class features while producing detailed heatmaps with better visual 
quality.


\vspace{-1mm}
\subsection{Measuring Visual Explanation Accuracy}
\label{sec:visualExplanationBenchmark}

\begin{figure}
  \centering
  \includegraphics[width=1\textwidth]{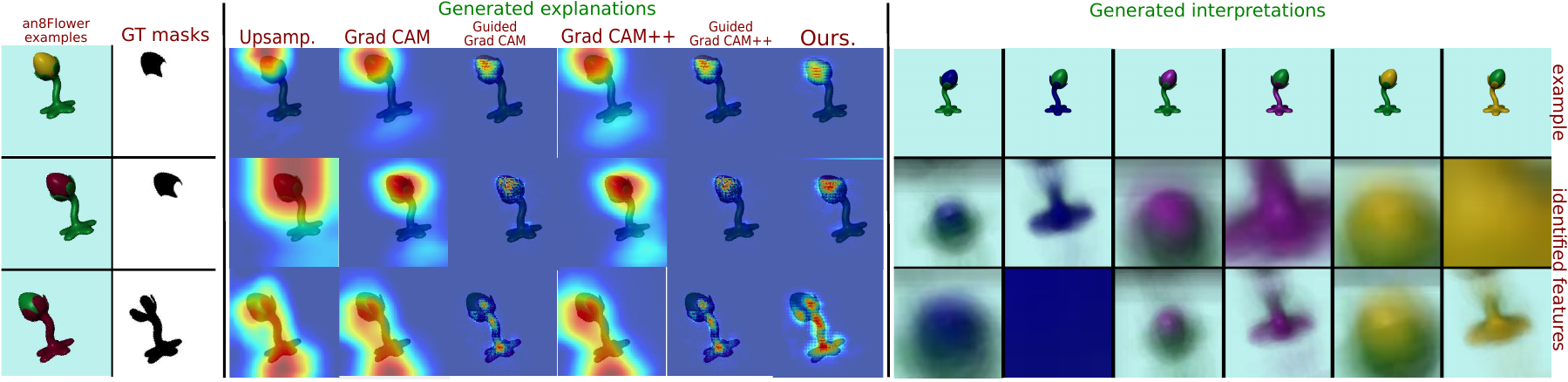}
  \vspace{-6mm}
  \caption{\small Left: Examples and GT-masks from the proposed \textit{an8FLower} dataset. Center: Comparison of generated visual explanations. Right: Examples of the generated visual interpretations.}
  
  \label{fig:syntheticExp}
  \vspace{-6mm}
\end{figure}

We generate two synthetic datasets, \textit{an8Flower-single-6c} and \textit{an8Flower-double-12c}, 
with 6 and 12 classes respectively.
In the former, a fixed single part of the object is allowed to change color.
This color defines the classes of interest. In the latter, a combination of 
color and the part on which it is located defines the discriminative feature. 
After defining these features, we generate masks that overlap 
with the discriminative regions (Fig.~\ref{fig:syntheticExp}\,(left)). 
Then, we threshold the heatmaps at given values and measure the pixel-level 
intersection over union (IoU) of a model explanation (produced by the method 
to be evaluated) w.r.t. these masks. 
We test a similar model as for the \MNIST ~dataset (Sec.~\ref{sec:importanceRelFeat}) 
trained on each variant of the \textit{an8Flower} dataset.
In Table 1 we report 5-fold cross-validation performance of the proposed 
feature selection method using three different means 
(Upsamp. Act. Maps, Deconv+GB~\cite{SpringenbergGuidedBack15} 
and ours heatmap variant) and other state-of-the-art methods 
to generate visual explanations.
%

We can notice in Fig.~\ref{fig:syntheticExp}\,(right) that our 
method effectively identifies the pre-defined discriminative regions regardless 
of whether they are related to color and/or shape.
Likewise, Fig.~\ref{fig:syntheticExp}\,(center) shows that
our explanations accurately highlight these features 
and that they have a better balance between 
level of detail and coverage than those produced by existing methods.
The quantitative results (Table~1) show that our method has a higher 
mean IoU of the discriminative features when compared to existing methods.
\jose{
However, it should be noted that, different from the compared methods, 
our method involves an additional process, i.e., feature selection via $\mu$-lasso, 
at training time. Moreover, for this process an additional parameter, i.e $\mu$, 
should be defined (Sec.~\ref{sec:identifyRelevanFeatures}).
}
Please refer to the supplementary material for more details and visualizations.

\jose{
Methods for model explanation/interpretation aim at providing users with insights 
on what a model has learned and why it makes specific predictions. Putting this together with the observations made in our experiments, there are two points that should be noted.
On the one hand, we believe that our objective evaluation should be complemented with simpler
user studies. This should ensure that the produced explanations are meaningful to the individuals they aim to serve.
On the other hand, our proposed evaluation protocol enables objective quantitative comparison of different methods for visual explanation. As such it is free of the weaknesses of exhaustive user studies and of the complexities that can arise when replicating them.
}

\subsection{Checking the Sanity of the Generated Visual Explanations}
\label{sec:sanityCheck}
\begin{figure}
  \centering
  \includegraphics[width=1\textwidth]{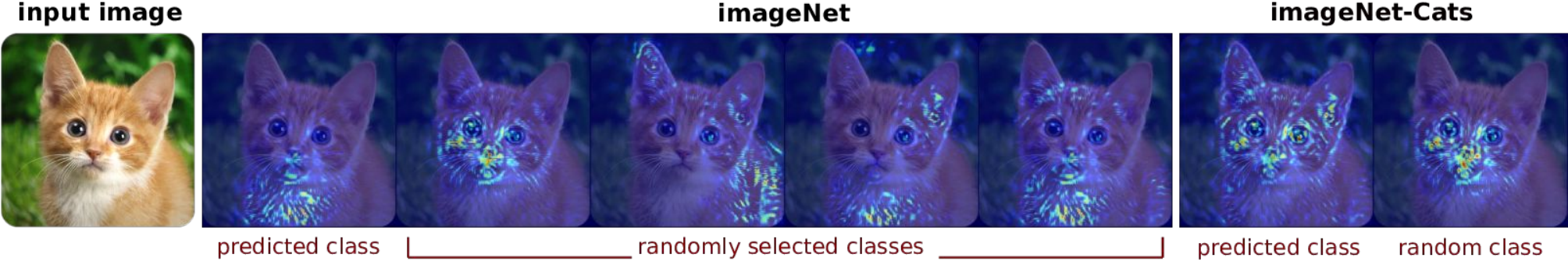}
  \vspace{-6mm}
  \caption{\small Sensitivity of the proposed method w.r.t. the predicted class. 
  Note how the generated explanation focuses on different regions of the input image (left) when explaining different classes.
  We see a similar trend on explanations generated from the models trained on the \imageNet~(center) and \imageNet-cats (right) datasets.
  }
  \label{fig:sanityCheck}
  \vspace{-6mm}
\end{figure}

\vspace{-2mm}
\jose{
Beyond the capability of generating accurate visual explanations, 
recent works (\cite{KindermansSanityNIPS17,SanityNIPS2018,NieZP18Sanity}) 
have stressed the importance of verifying that the generated explanations 
are indeed relevant to the model and the classes being explained.
Towards this goal, we run a similar experiment to that conducted in \cite{NieZP18Sanity} 
where the visual explanation produced for a \textit{predicted class}
of a given model after observing a given image is compared against those when a 
\textit{different class} is considered when generating the explanation.
A good explanation should be sensible to the class, and thus generate different visualizations.
In Fig.~\ref{fig:sanityCheck} we show qualitative results obtained by running this 
experiment on our models trained on the \ILSVRC'12~(\cite{ILSVRC15}) and ~\imageNet-cats datasets.
}

\jose{As can be noted in Fig.~\ref{fig:sanityCheck}, the explanation generated for the predicted class, i.e., \textit{'cat'/'tabby'}, 
focuses on different regions than those generated for randomly selected classes.
This is more remarkable for the case of the \imageNet-cats model, which can be considered 
a fine-grained classification task.  
In this setting, when changing towards a random class, i.e., \textit{'jaguar, panther, Panthera onca, Felis onca'}, 
the generated explanations only highlight the features that are common between the random class 
and the \textit{'tabby'} class depicted in the image.
In their work, \cite{NieZP18Sanity} and \cite{SanityNIPS2018} found that explanations 
from DeconvNet and Guided-Backpropagation methods are not performing well in this respect, yielding visualizations that are not determined by the predicted 
class, but by the filters of the first layer and the edge-like structures in the 
input images. Although our method relies on DeconvNet and Guided-Backpropagation, 
our explanations go beyond regions with prominent gradients 
- see Fig.~\ref{fig:teaser}, \ref{fig:visualExplanationImages} \& \ref{fig:syntheticExp}. In fact, in classes where color is a discriminative feature, 
uniform regions are highlighted. 
This different result can be understood since, in our method,  DeconvNet with Guided-Backpropagation is merely used as a means to 
highlight the image regions that justify the identified relevant features, not the 
predicted classes themselves. 
If a better, more robust or principled
visualization method is proposed in the future by the community, we could use that as well.
}

\section{Conclusion}
\label{sec:conclusion}
\vspace{-2mm}
We propose a method to enrich the prediction made by 
\DNNs~by indicating the visual features that 
contributed to such prediction.
Our method identifies features encoded
by the network that are relevant for the task 
addressed by the \DNN. 
It allows \textit{interpretation} of these features 
by the generation of average feature-wise visualizations.
In addition, we proposed a method to attenuate 
the artifacts introduced by strided operations in 
visualizations made by Deconvnet-based methods.
This empowers our method with richer visual feedback 
with pixel-level precision without requiring additional
annotations for supervision.
Finally, we have proposed a novel dataset designed
for the objective evaluation of methods for explanation of \DNNs.


%
\subsubsection*{Acknowledgments} This work was supported by the \small{FWO SBO} project Omnidrone, 
the \small{VLAIO} R\&D-project \small{SPOTT} 
, the KU Leuven \small{PDM} Grant \small{PDM/16/131}, 
and a \small{NVIDIA GPU} grant.
%

\small
\bibliographystyle{iclr2019_conference}
\bibliography{iclr2019_conference}


\newpage

\section{Supplementary Material}

This section constitutes supplementary material.
As such, this document is organized in six parts. 
In Section~\ref{sec:implementDetails}, we provide implementation 
details of the proposed method and experiments presented in 
the original manuscript.
In Section~\ref{sec:relevantFeatures}, we provide a further 
quantitative analysis on the importance of the identified
relevant features. 
In Section~\ref{sec:AD}, we provide additional 
details regarding the generation of \textit{an8Flower}, a 
dataset specially designed for evaluating methods for visual 
explanation.
In Section~\ref{sec:visualInterpretation}, 
we provide extended examples on the average images used 
for interpretation. 
Similarly, in Section~\ref{sec:visualExplanation}, 
we provide additional examples of visual explanations 
provided by our method. 
Finally, we concluded this document
in Section~\ref{sec:qualityComparison} by performing 
a qualitative comparison in order to display the advantages 
of the proposed method over existing work.

\subsection{Implementation Details}
\label{sec:implementDetails}

We use in our experiments the pre-trained models provided as part 
of the MatconvNet framework~\cite{MatConvNet} for both the \MNIST 
and ImageNet datasets. For the \Fashion144k dataset we use the \VGG-\F-based 
model (\textit{Finetunned with image-based pooling}) released by the authors 
from \cite{wangGeo17}.
For \MNIST, we employ a network composed by 8 layers in total, 
five of them are convolutional , two are fully connected. 
The last one is a softmax layer.
For the full \imageNet set, we employ a \VGG-\F~\cite{Chatfield14} model
which is composed of 21 layers, from these 15 are convolutional followed
by five fully connected. The last one is a softmax layer.
Finally, for the case of the \imageNet-Cats subset
we finetune the \VGG-\F model trained on the full \imageNet set.

\subsection{Importance of the Identified Features}
\label{sec:relevantFeatures}

\begin{figure}[h!]
\centering
\includegraphics[width=0.3\textwidth]{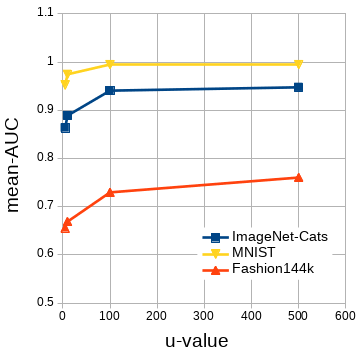}
\caption{
Classification based on the identified relevant 
features. 
We present the mean area under the ROC curve (mean-AUC) 
for all the classes of interest in a given dataset.
Note that only a small amount of features, i.e., low $\mu$, 
is required to produce a good reconstruction. 
}
\label{fig:classReconstruction}
\end{figure}

In order to verify the relevance of the identified 
features, i.e., how well the features encode information
from the classes of interest, we measure the level to 
which they are able to "reconstruct" each of the 
classes of interest.
Towards this goal, we compute the mean area under 
the ROC curve (mean-AUC) for all the classes of interest
in a given dataset.
In Figure~\ref{fig:classReconstruction}, we report 
performance over different $\mu$ values.
It can be noted that already with a low amount of selected 
features, i.e.. a low value of $\mu$, we can already encode, 
properly, visual characteristics of the classes of interest.
This shows that the identified features hold strong potential 
as visual means for explanation.


\subsection{an8Flower Dataset}
\label{sec:AD}

We release an8Flower, a dataset and evaluation protocol specifically designed for evaluating the performance of methods for model explanation.
We generate this dataset by taking as starting point the eggplant model~\footnote{http://www.anim8or.com/learn/tutorials/eggplant/index.html} publicly released with the Anim8or~\footnote{http://www.anim8or.com/} 3D modeling software. Then, we introduce into this model the discriminative features to define each of the classes of interest. For each class of interest, we rotate the object 360 degrees and render/save 40 frames at different viewpoints with a size of 300x300 pixels.

\begin{figure*}
\centering
\includegraphics[width=1\textwidth]{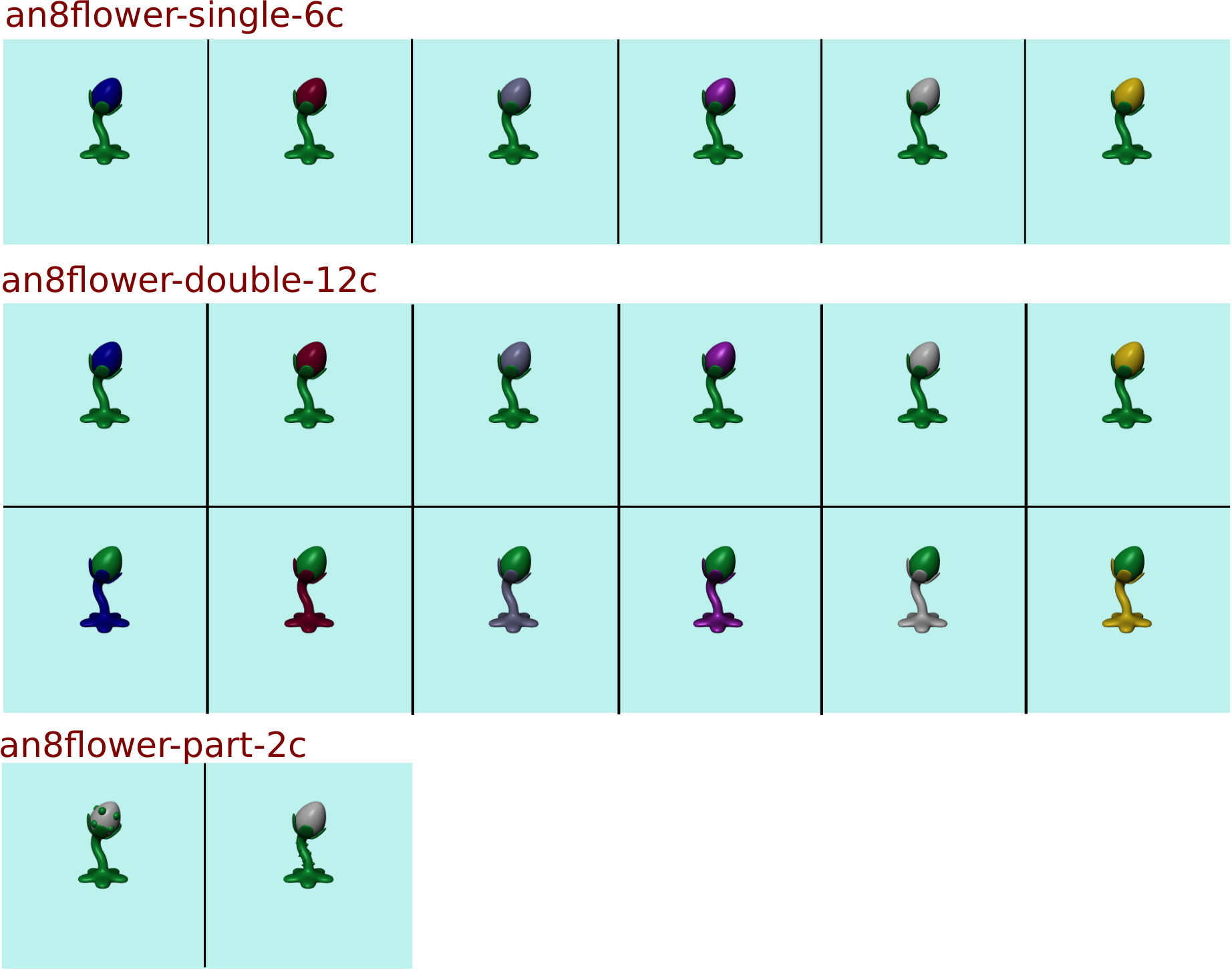}

\caption{
Rendered images from three variant of the proposed \textbf{an8Flower} dataset. \textit{an8Flower-single-6c}
contains six different flower colors, \textit{an8Flower-double-12c} has six more
classes, the same six colors but focused on the stem part. \textit{an8Flower-part-2c} consists of two
classes defined by the occurrence of balls on the flower or thorns on the stem part.
}
\label{fig:AD_examples}
\end{figure*}

\begin{figure*}
\centering
\includegraphics[width=1\textwidth]{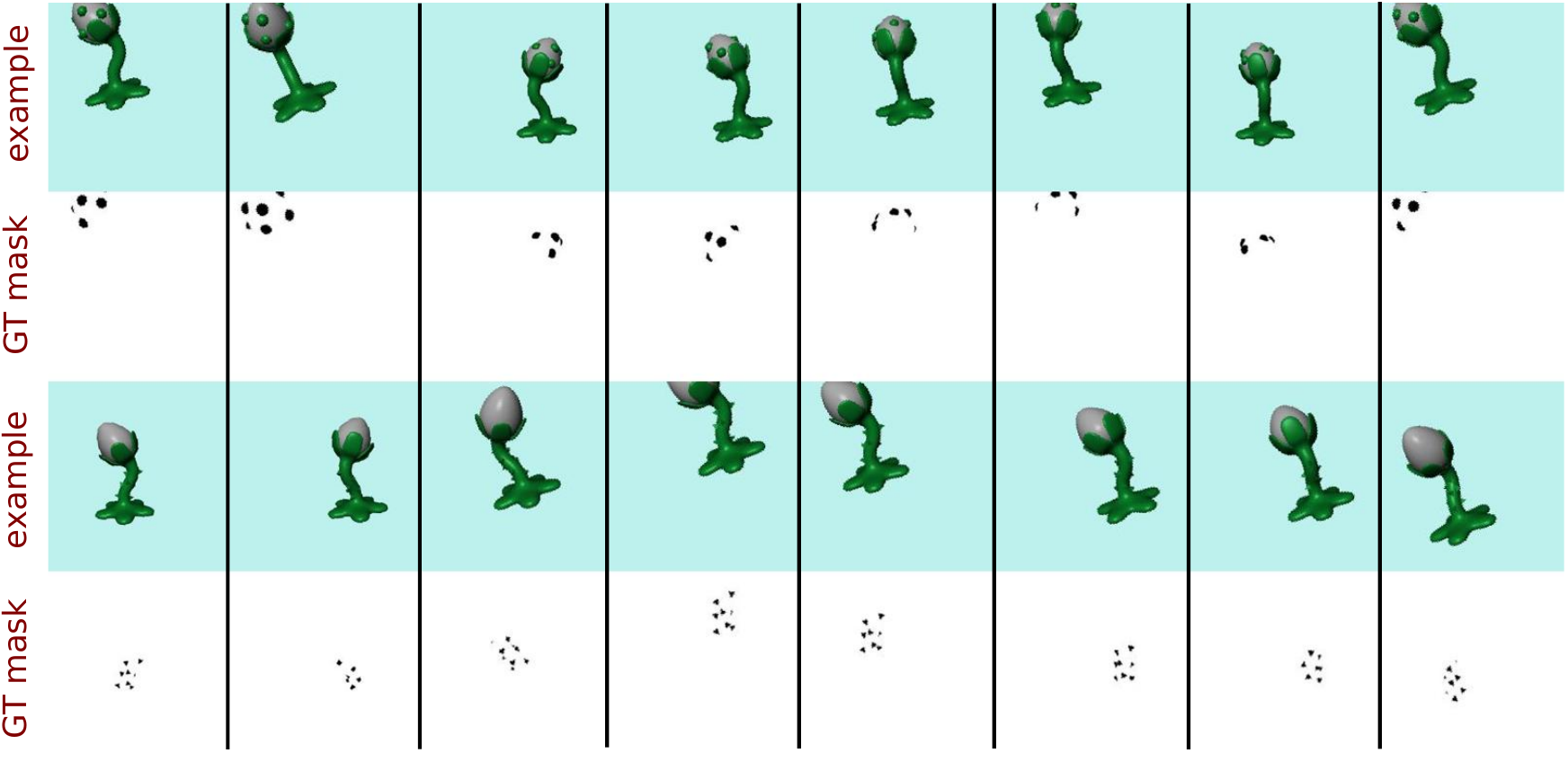}

\caption{
Random selection of examples from the \textit{an8Flower-part-2c} with their corresponding ground-truth masks.
}
\label{fig:AD_dataArg_part}
\end{figure*}

\begin{figure*}
\centering
\includegraphics[width=1\textwidth]{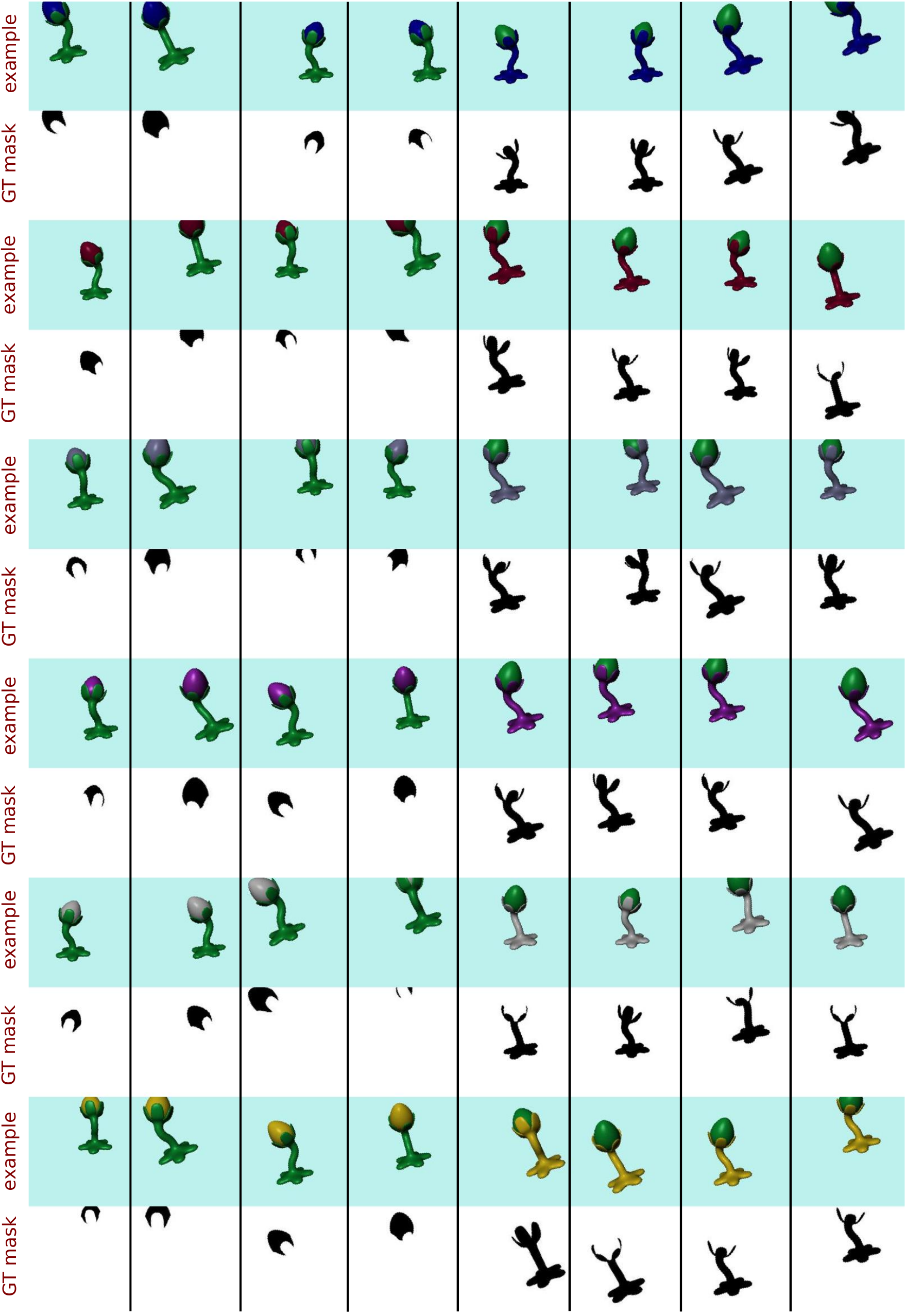}

\caption{
Random selection of examples from the \textit{an8Flower-single-6c} and \textit{an8Flower-double-12c} with their corresponding ground-truth masks
}
\label{fig:AD_dataArg_color}
\end{figure*}

Afterwards, in order to increase variation and the amount of data, we apply the following data augmentation procedure on each image frame. For each of the original 300x300 rendered frames, we crop each image five times at four corners and center respectively with the size of 250x250. Then, each cropped image is rotated five angles: 5,~10,~15,~20 and 25 degrees. In the end, the data augmentation produces a total of 1000 example images per class.
Figure~\ref{fig:AD_examples} shows rendered images of the classes of interest considered in three variants of this dataset, i.e., \textit{an8Flower-single-6c}, \textit{an8Flower-double-12c} and \textit{an8Flower-part-2c}.
Figure~\ref{fig:AD_dataArg_part} and Figure ~\ref{fig:AD_dataArg_color} displays some examples of the augmented data used in training/testing and its corresponding mask. 

\subsection{Visual Interpretation}
\label{sec:visualInterpretation}

In this section, we provide extended examples 
of the average images used by our method as 
means for visual interpretation 
\textit{(Section \ref{sec:importanceRelFeat} of the original manuscript)}. 
Moreover, we provide a visual comparison with their 
counterparts generated by existing methods. 
More precisely, up-scaled activations maps 
from convolutional layers~\cite{netdissect2017,Zhou15Parts} 
and heatmaps generated from deconvnets with 
guided backpropagation~\cite{gruen16,SpringenbergGuidedBack15,ZeilerDeconv14}.

\begin{figure*}
\includegraphics[width=1\textwidth]{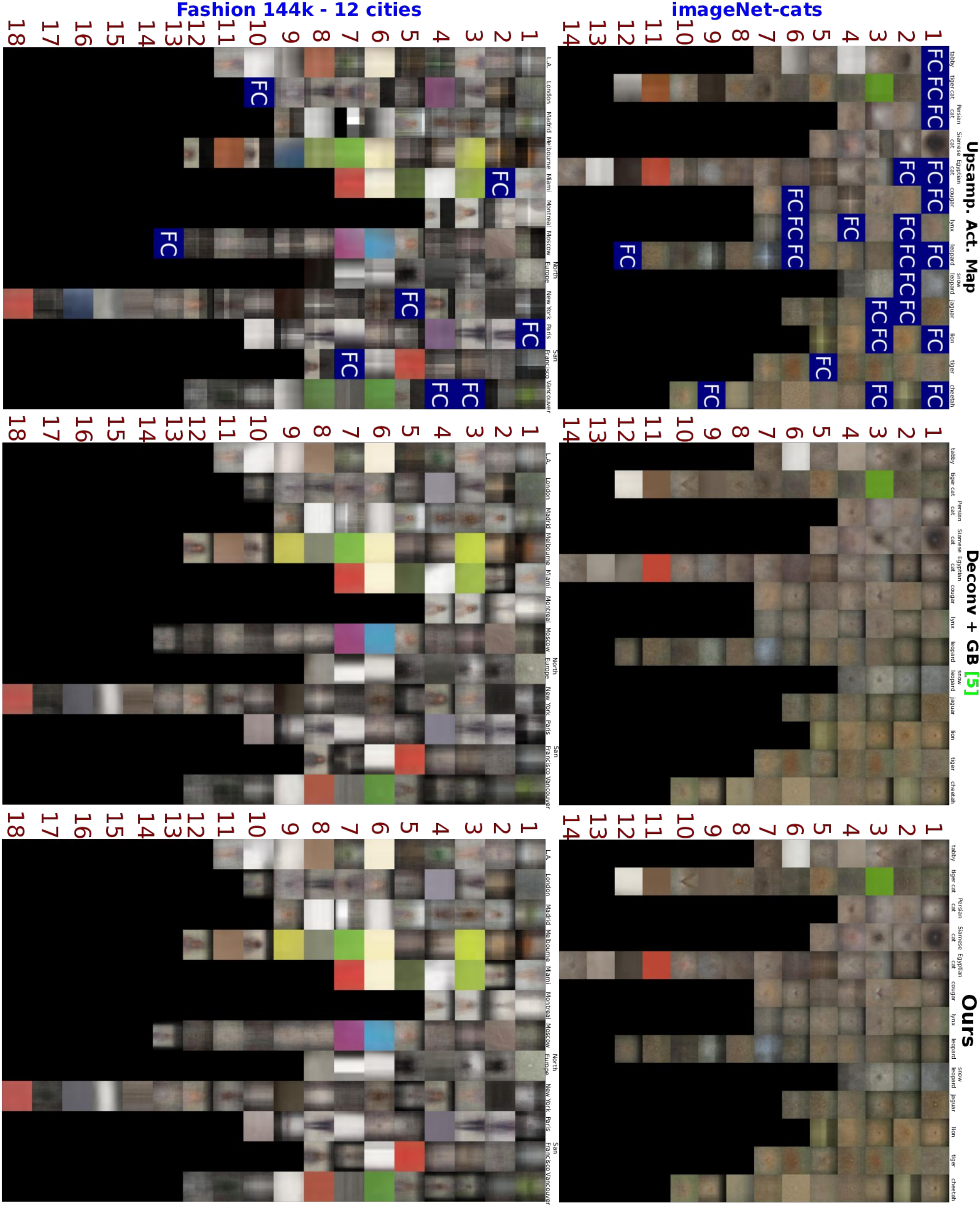}
\caption{
\textbf{Interpretation} (Average) visualizations for the identified relevant features for the \textbf{imageNet-Cats}~\cite{ILSVRC15} and a subset of the \textbf{Fashion144k}~\cite{SimoSerra15} dataset. For each class on each dataset, average features are sorted by decreasing relevance in the class they encode.
Average images are generated by either considering: up-scaled activation maps, heatmaps from methods based on deconvnet with guided-backpropagation (deconv+GB)~\cite{SpringenbergGuidedBack15}, or our method.
}
\label{fig:avgImg144Cats}
\end{figure*}

\begin{figure*}
\hspace*{-6mm}\includegraphics[width=1.08\textwidth]{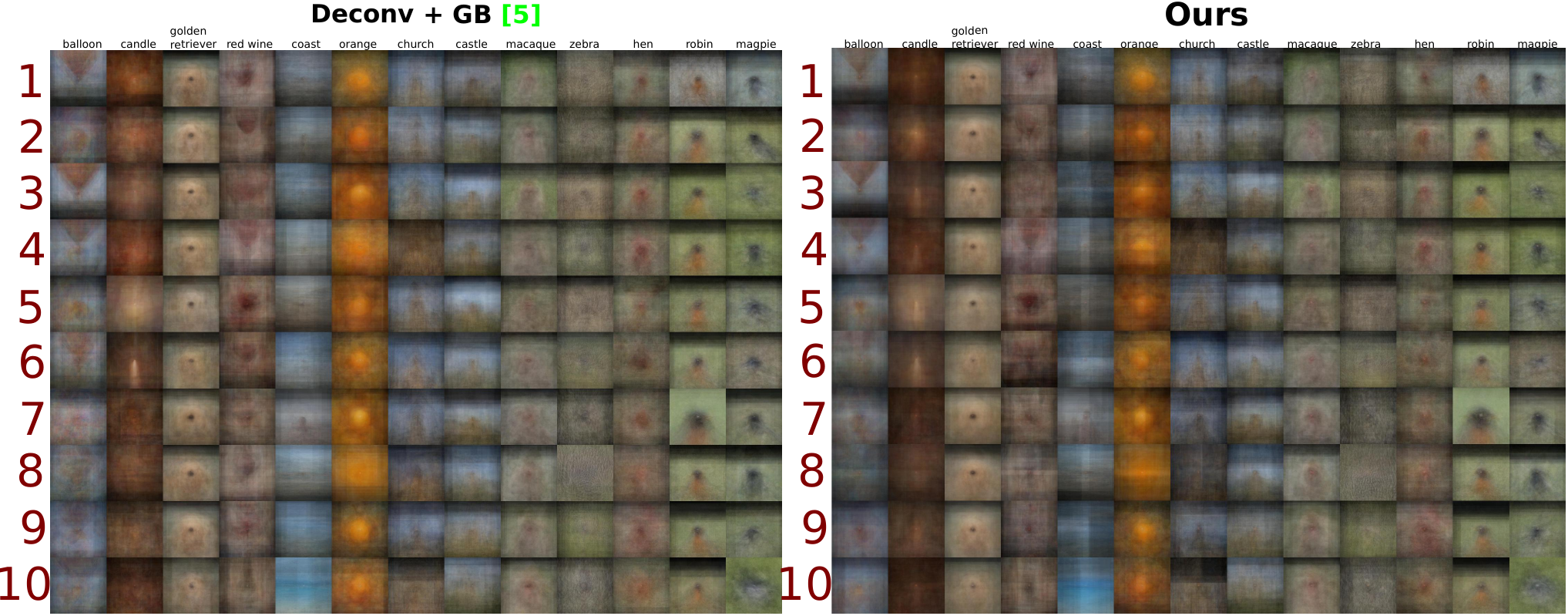}
\caption{
\textbf{Interpretation} (Average)  visualizations for the identified relevant features for a subset of classes from the \textbf{imageNet}~\cite{ILSVRC15} dataset. 
Average images are generated by either considering heatmaps from methods based on deconvnet with guided-backpropagation (deconv+GB)~\cite{SpringenbergGuidedBack15} (left) and our method (right).
}
\label{fig:imageNetAvgImg}

\hspace*{-6mm}\includegraphics[width=1.08\textwidth]{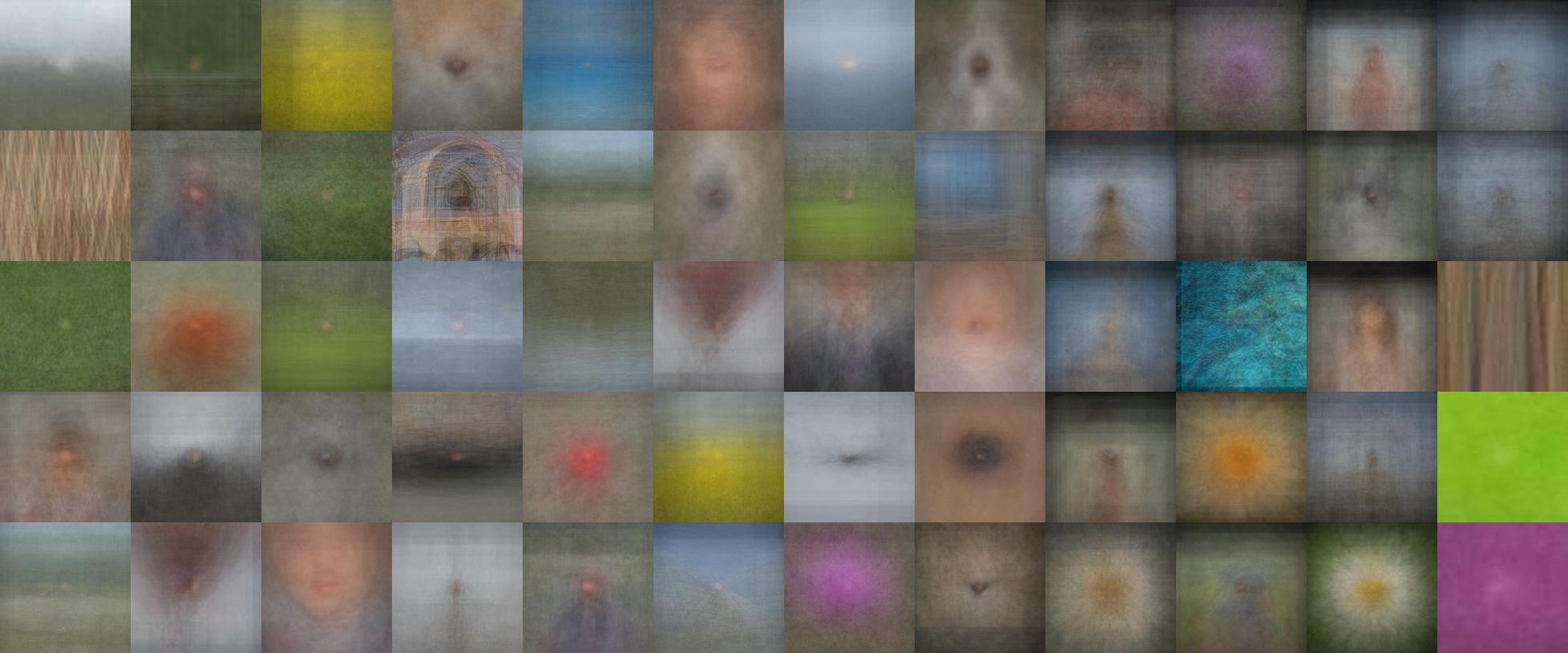}
\caption{
\textbf{Interpretation} (Average)  visualizations produced by our method for the identified relevant features for the \textbf{imageNet}~\cite{ILSVRC15} dataset.
}
\label{fig:imageNetAvgImgMix}
\end{figure*}

In Figure~\ref{fig:avgImg144Cats}, we show 
average images for the imageNet-Cats~\cite{ILSVRC15} 
subset, and the Fashion144k~\cite{SimoSerra15} 
datasets, respectively.
For each class on each dataset, we show the 
average image of each identified feature sorted, 
from top to bottom, based on their relevance for 
its corresponding class.
In a similar fashion, in Figure~\ref{fig:imageNetAvgImg}, 
we show a visualization for the displayed 
subset of classes \textit{(Figure 5 (center) in 
the original manuscript)} from the imageNet dataset.
Given that for the imageNet dataset most of the 
identified features come from fully connected (FC) 
layers, no up-scaled response map visualization is 
possible for them.
Therefore, we opted not to display average 
visualizations based on up-scaled activation maps.
Finally, in Figure~\ref{fig:imageNetAvgImgMix}, we show 
average visualizations for additional classes from
imageNet.

In Figure~\ref{fig:avgImg144Cats} we can notice 
that already within few top relevant 
features per class, semantic concepts start to 
appear.
In addition, we see the same trends observed
in the original paper. On the one hand, some 
features encode class-specific  properties, 
e.g. nose shape, fur pattern, for the case of 
the imageNet-Cats, or hair color/length, leg 
dressing, or clothing color for the case of 
Fashion 144k.
On the other hand, some features encode 
properties related to the background/context
in which the images are captured, e.g. wall/road 
colors, vegetation, etc.
We can notice that for the case of up-sampled 
activation maps, there are identified features
which originate in Fully Connected Layers (FC).
For this type of layer, the up-scaling process
is not applicable.

When comparing the average visualizations 
generated by the different methods, it is 
noticeable that the proposed method produces 
sharper visualizations than those generated from 
up-scaled activation maps.
Moreover, our visualizations are still sharper 
than those produced by state-of-the-art 
deconvnet-based methods with 
guided-backpropagation~\cite{SpringenbergGuidedBack15}.
Therefore, enhancing the interpretation capabilities 
of the proposed method.

\subsection{Visual Explanation}
\label{sec:visualExplanation}
\begin{figure*}
\centering
\includegraphics[width=1\textwidth]{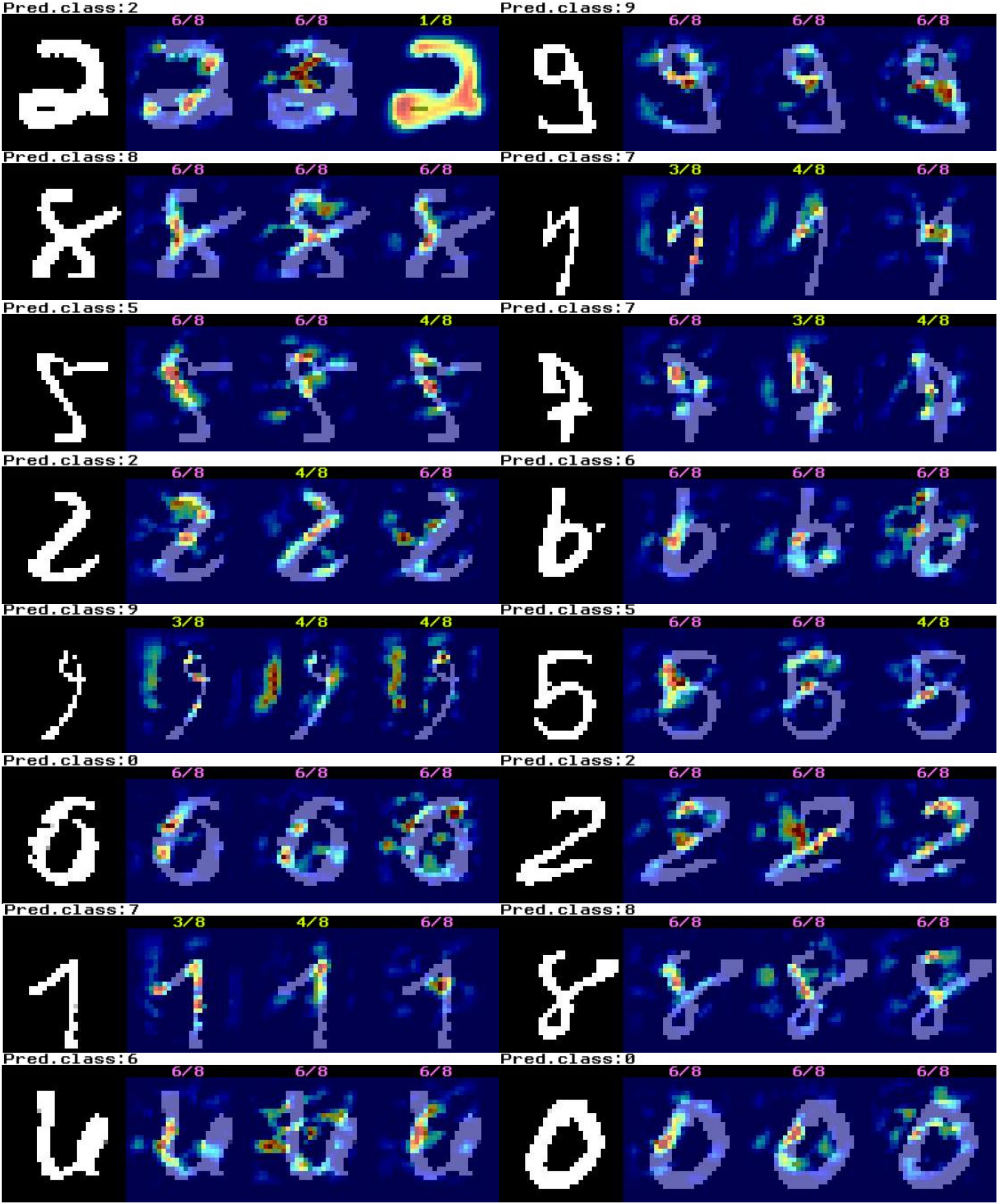}

\caption{ \textbf{Generated visual explanations} from the \textbf{MNIST} dataset.
  We accompany the predicted class label with our 
  heatmaps indicating the pixel locations, associated to the features, 
  that contributed to the prediction.
  On top of each heatmap we indicate the number of the layer where the features come
  from. The layer type is color-coded, i.e., convolutional (green) and fully connected (pink).
}
\label{fig:visualExplanationMNIST}
\end{figure*}

\begin{figure*}
\centering
\includegraphics[width=1\textwidth]{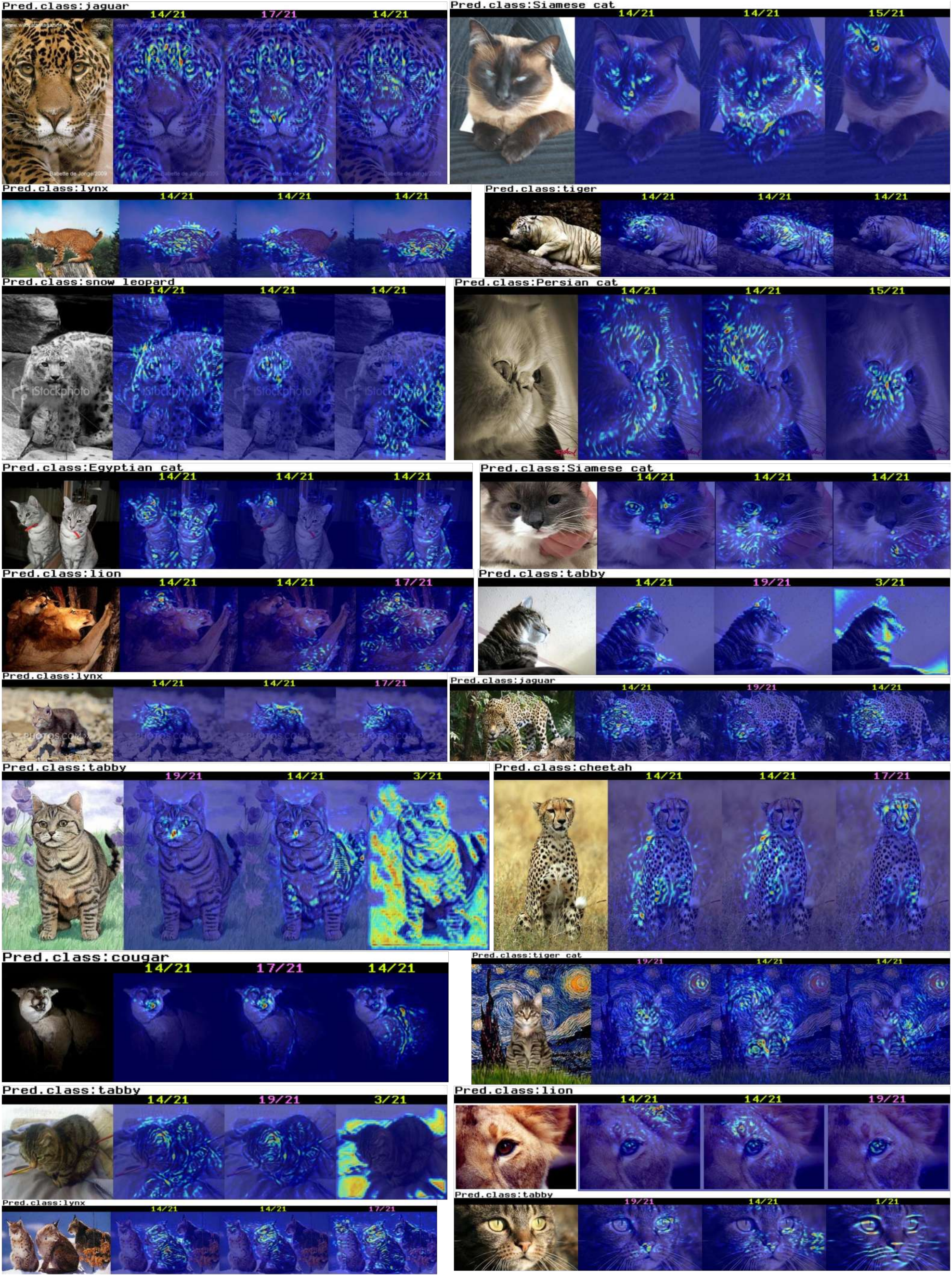}

\caption{
 \textbf{Generated visual explanations} from the \textbf{imageNet-Cats}~\cite{ILSVRC15} subset.
  We accompany the predicted class label with our 
  heatmaps indicating the pixel locations, associated to the features, 
  that contributed to the prediction.
  On top of each heatmap we indicate the number of the layer where the features come
  from. The layer type is color-coded, i.e., convolutional (green) and fully connected (pink).}
\label{fig:visualExplanationImageNetCats}
\end{figure*}

\begin{figure*}
\centering
\includegraphics[width=1\textwidth]{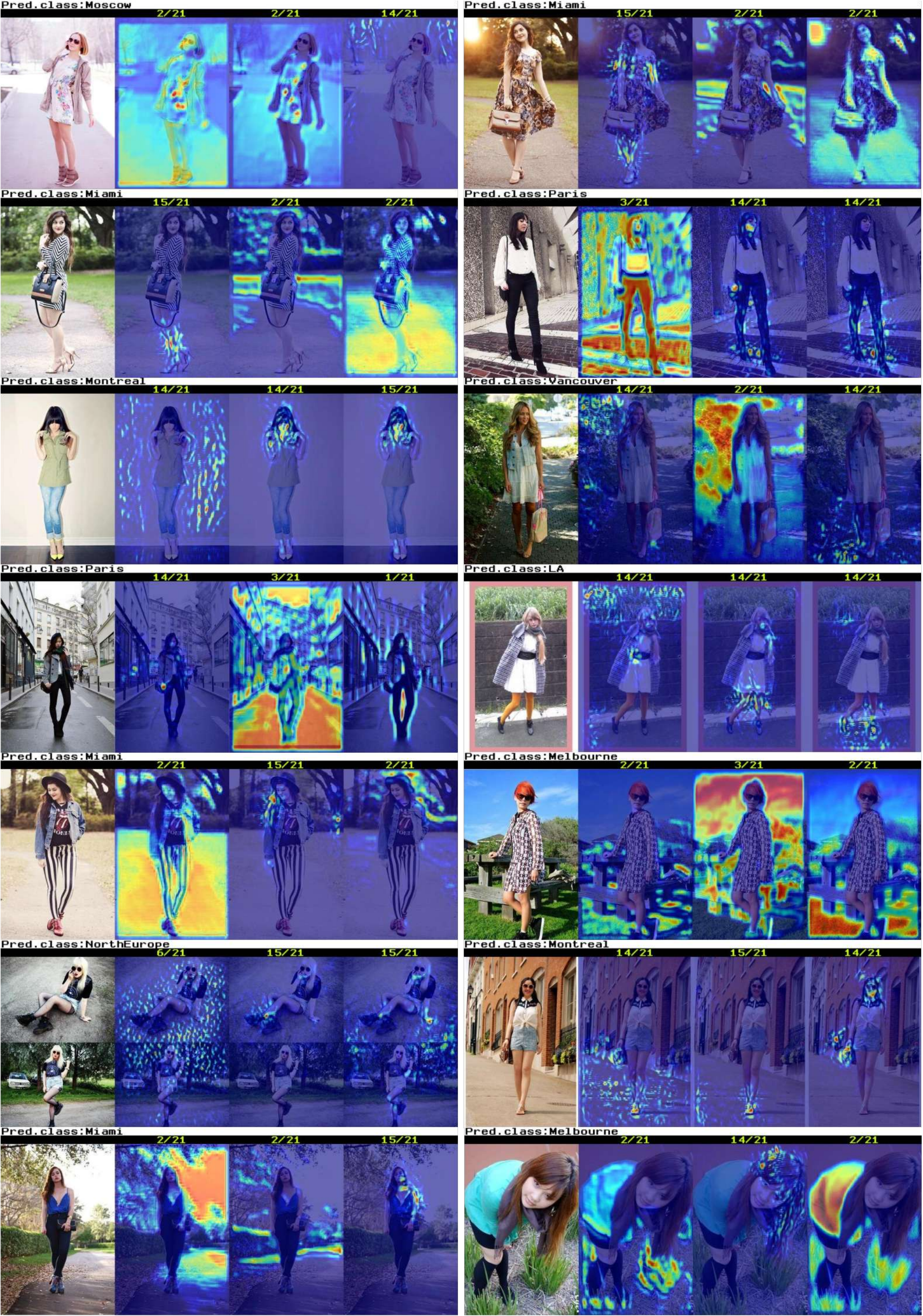}

\caption{
\small \textbf{Generated visual explanations} from the \textbf{Fashion144k} dataset~\cite{SimoSerra15}.
  We accompany the predicted class label with our 
  heatmaps indicating the pixel locations, associated to the features, 
  that contributed to the prediction.
  On top of each heatmap we indicate the number of the layer where the features come
  from. The layer type is color-coded, i.e., convolutional (green) and fully connected (pink).
}
\label{fig:visualExplanationFashion144k}
\end{figure*}

In Figures~\ref{fig:visualExplanationMNIST}, \ref{fig:visualExplanationImageNetCats}, \ref{fig:visualExplanationFashion144k}, \ref{fig:AD_dataArg_color}, \ref{fig:AD_dataArg_part} 
we extend the visual explanation results presented in the original manuscript.
\textit{(Figures 1, 6, 7 and 9 in the original manuscript)}

In the visual explanations generated by our method 
we accompany the predicted class label with our 
heatmaps indicating the pixel locations, associated 
to the features, that contributed to the prediction.
In line with the original manuscript, on top of each 
heatmap we indicate the number of the layer where 
the features come from. 

\begin{figure*}
\centering
\includegraphics[width=1\textwidth]{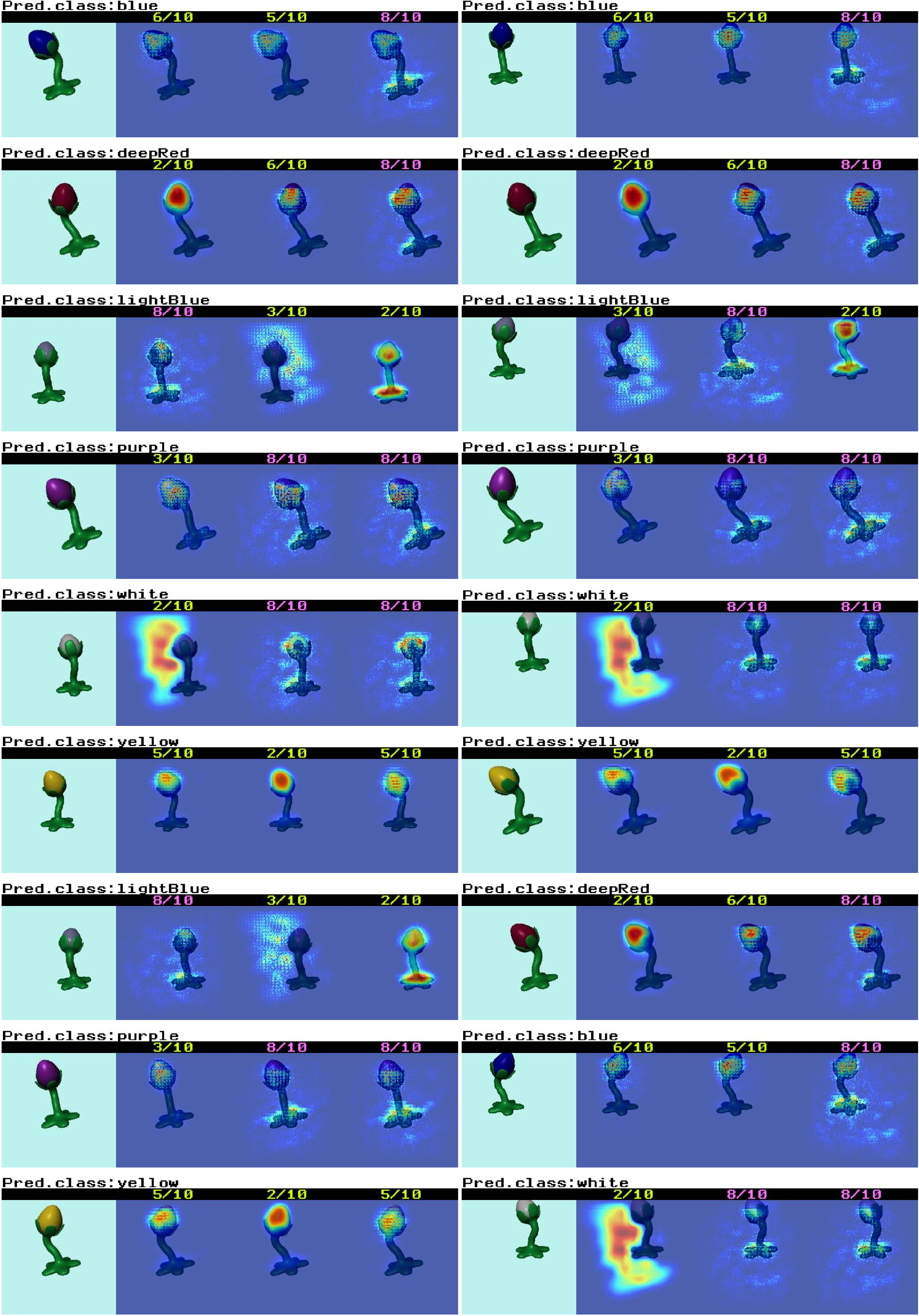}

\caption{
\small \textbf{Generated visual explanations} from our proposed \textbf{An8Flower-single-6c} dataset.
  We accompany the predicted class label with our 
  heatmaps indicating the pixel locations, associated to the features, 
  that contributed to the prediction.
  On top of each heatmap we indicate the number of the layer where the features come
  from. The layer type is color-coded, i.e., convolutional (green) and fully connected (pink).
}
\label{fig:visualExplanationAn8SingleColor}
\end{figure*}


\begin{figure*}
\centering
\includegraphics[width=0.8\textwidth]{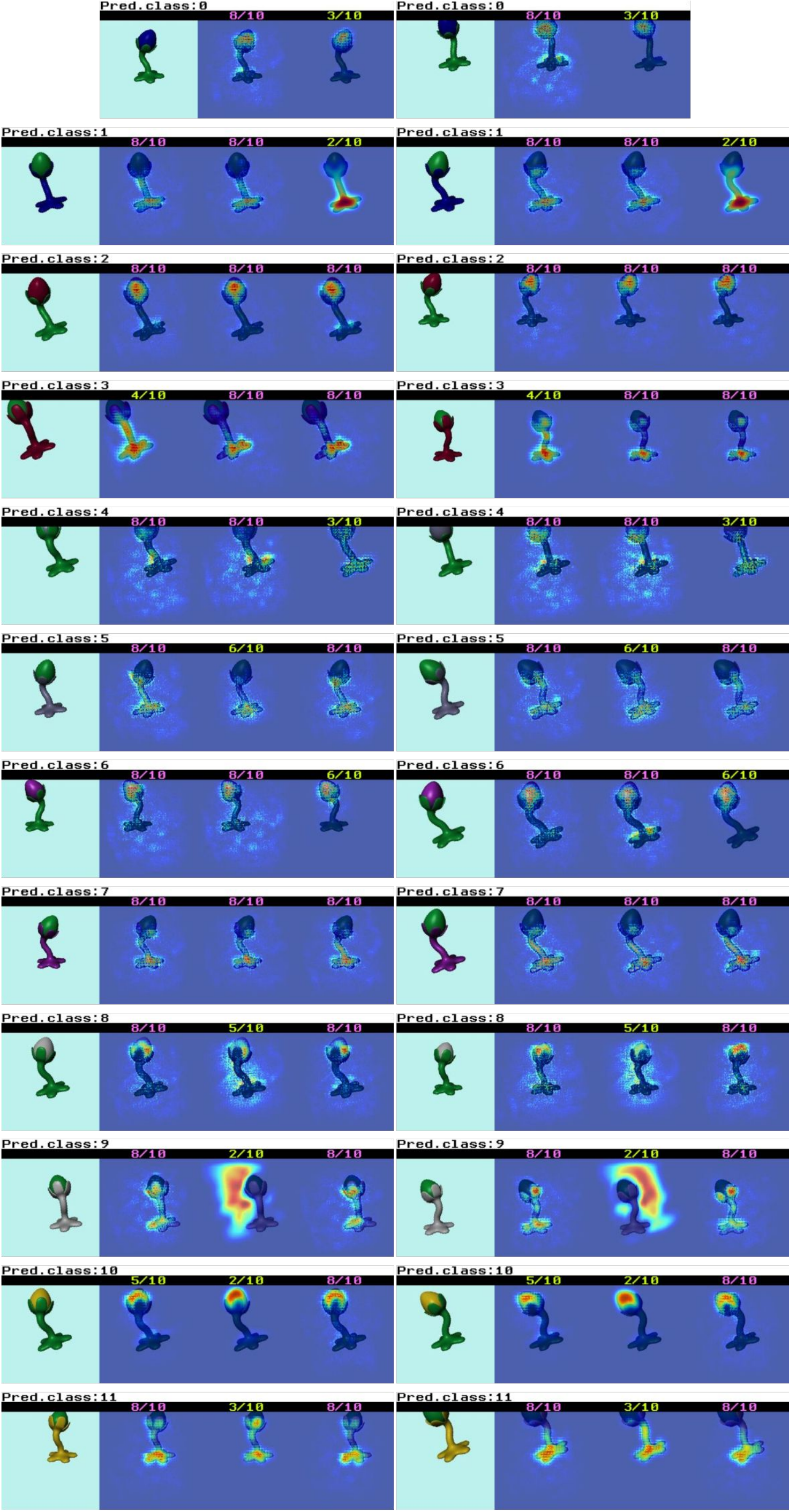}

\caption{
\small \textbf{Generated visual explanations} from our proposed \textbf{An8Flower-double-12c} dataset.
  We accompany the predicted class label with our 
  heatmaps indicating the pixel locations, associated to the features, 
  that contributed to the prediction.
  On top of each heatmap we indicate the number of the layer where the features come
  from. The layer type is color-coded, i.e., convolutional (green) and fully connected (pink).
}
\label{fig:visualExplanationAn8DoubleColor}
\end{figure*}


\begin{figure*}
\centering
\includegraphics[width=1\textwidth]{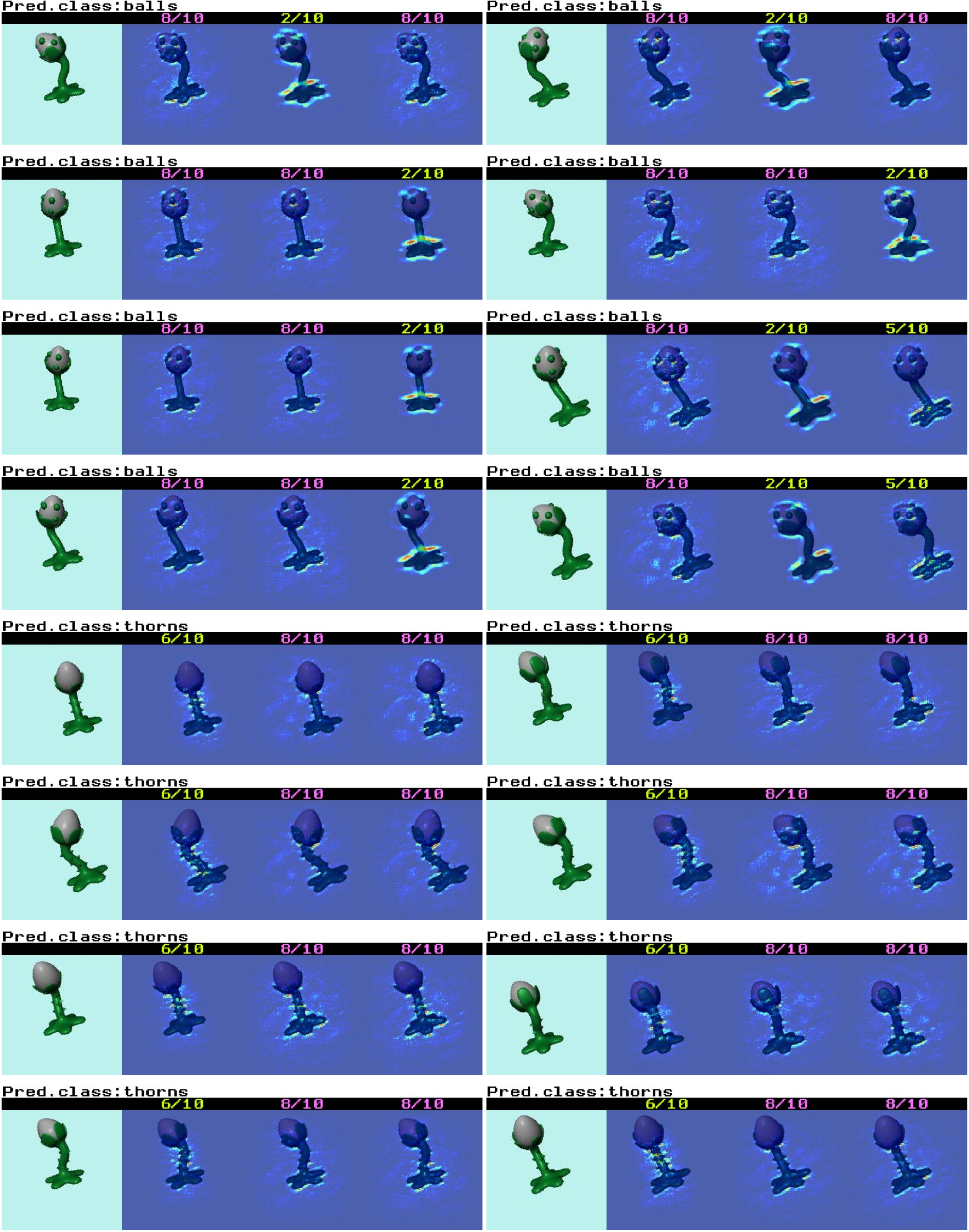}

\caption{
\small \textbf{Generated visual explanations} from our proposed \textbf{An8Flower-part-2c} dataset.
  We accompany the predicted class label with our 
  heatmaps indicating the pixel locations, associated to the features, 
  that contributed to the prediction.
  On top of each heatmap we indicate the number of the layer where the features come
  from. The layer type is color-coded, i.e., convolutional (green) and fully connected (pink).
}
\label{fig:visualExplanationAn8SinglePart}
\end{figure*}


\subsection{Visual Quality Comparison}
\label{sec:qualityComparison}

\begin{figure*}
\centering
\includegraphics[width=1\textwidth]{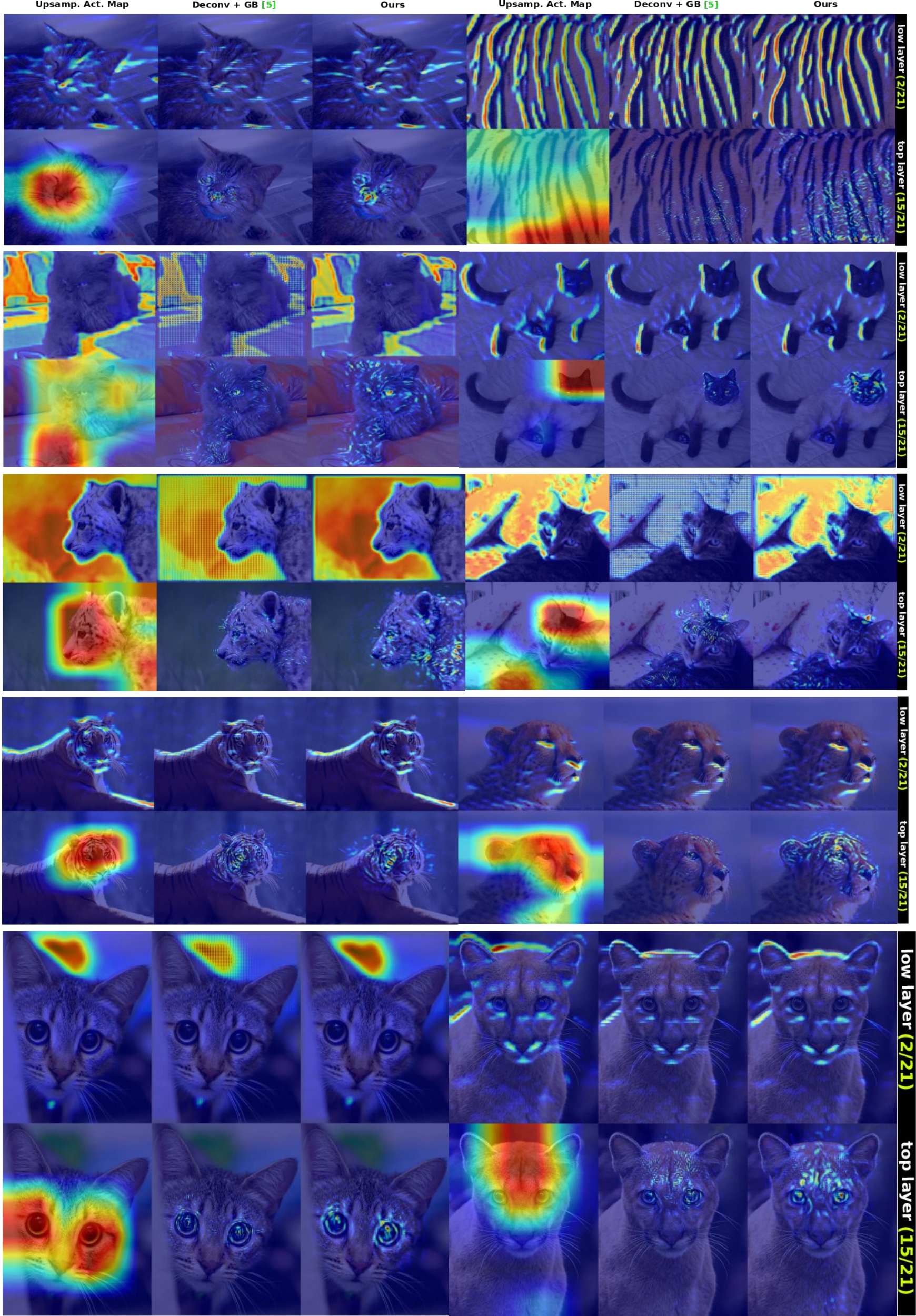}

\caption{
\textbf{Visual quality comparison} for visualizations generated from the \textbf{imageNet-Cats} subset~\cite{ILSVRC15}. Note how our heatmaps attenuate the grid-like artifacts introduced by deconvnet-based methods at lower layers. Likewise, our method is able to produce a more detailed visual feedback than upsampled activation maps.
}
\label{fig:visualQualityImageNetCats1}
\end{figure*}

\begin{figure*}
\centering
\includegraphics[width=1\textwidth]{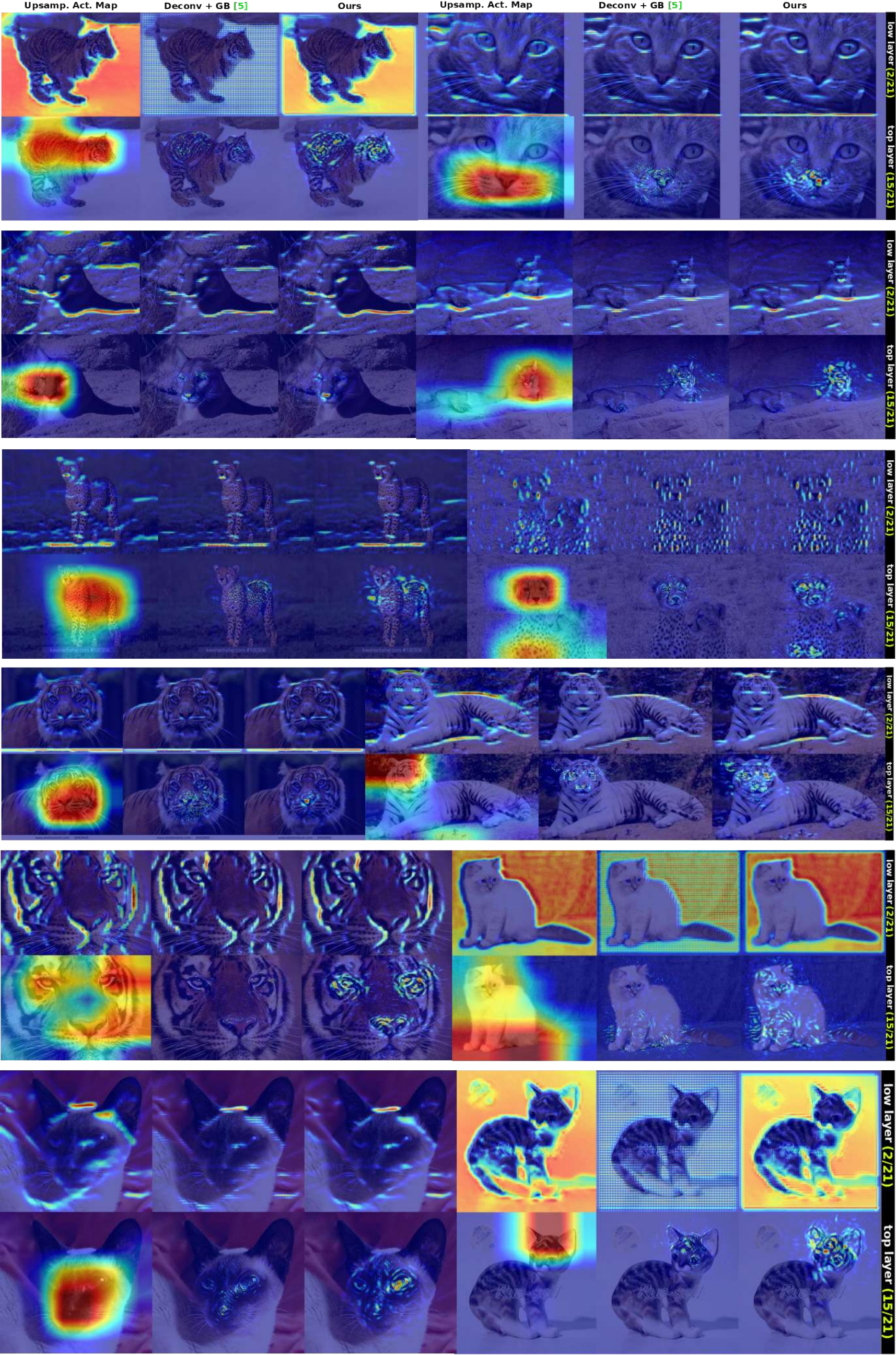}

\caption{
\textbf{Visual quality comparison} for visualizations generated from the \textbf{imageNet-Cats} subset~\cite{ILSVRC15}. Note how our heatmaps attenuate the grid-like artifacts introduced by deconvnet-based methods at lower layers. Likewise, our method is able to produce a more detailed visual feedback than upsampled activation maps.
}
\label{fig:visualQualityImageNetCats2}
\end{figure*}

\begin{figure*}
\centering
\includegraphics[width=1\textwidth]{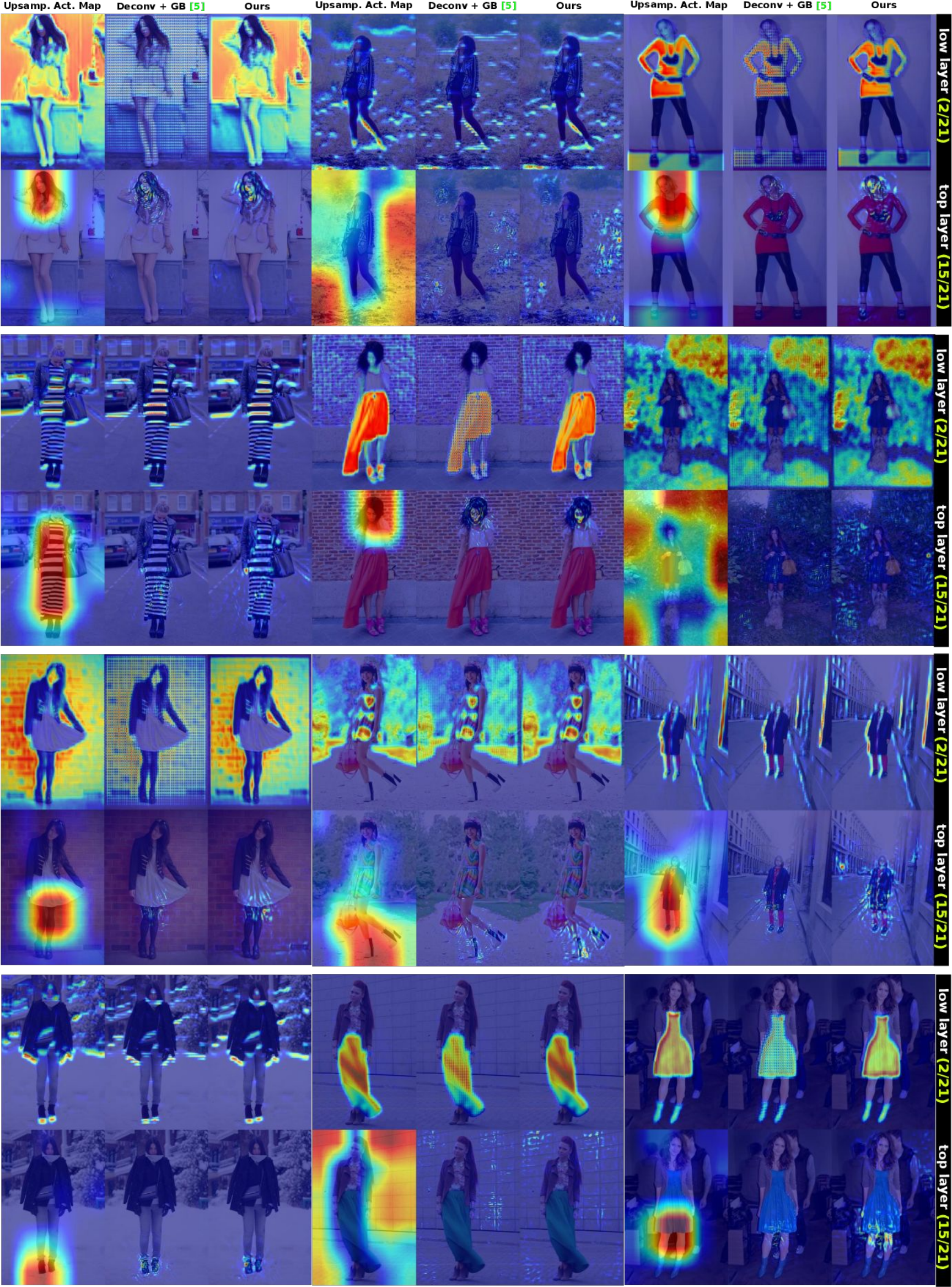}

\caption{
\textbf{Visual quality comparison} for visualizations generated from the \textbf{Fashion114k} dataset~\cite{SimoSerra15}. Note how our heatmaps attenuate the grid-like artifacts introduced by deconvnet-based methods at lower layers. Likewise, our method is able to produce a more detailed visual feedback than upsampled activation maps.
}
\label{fig:visualQualityFashion144k1}
\end{figure*}

\begin{figure*}
\centering
\includegraphics[width=1\textwidth]{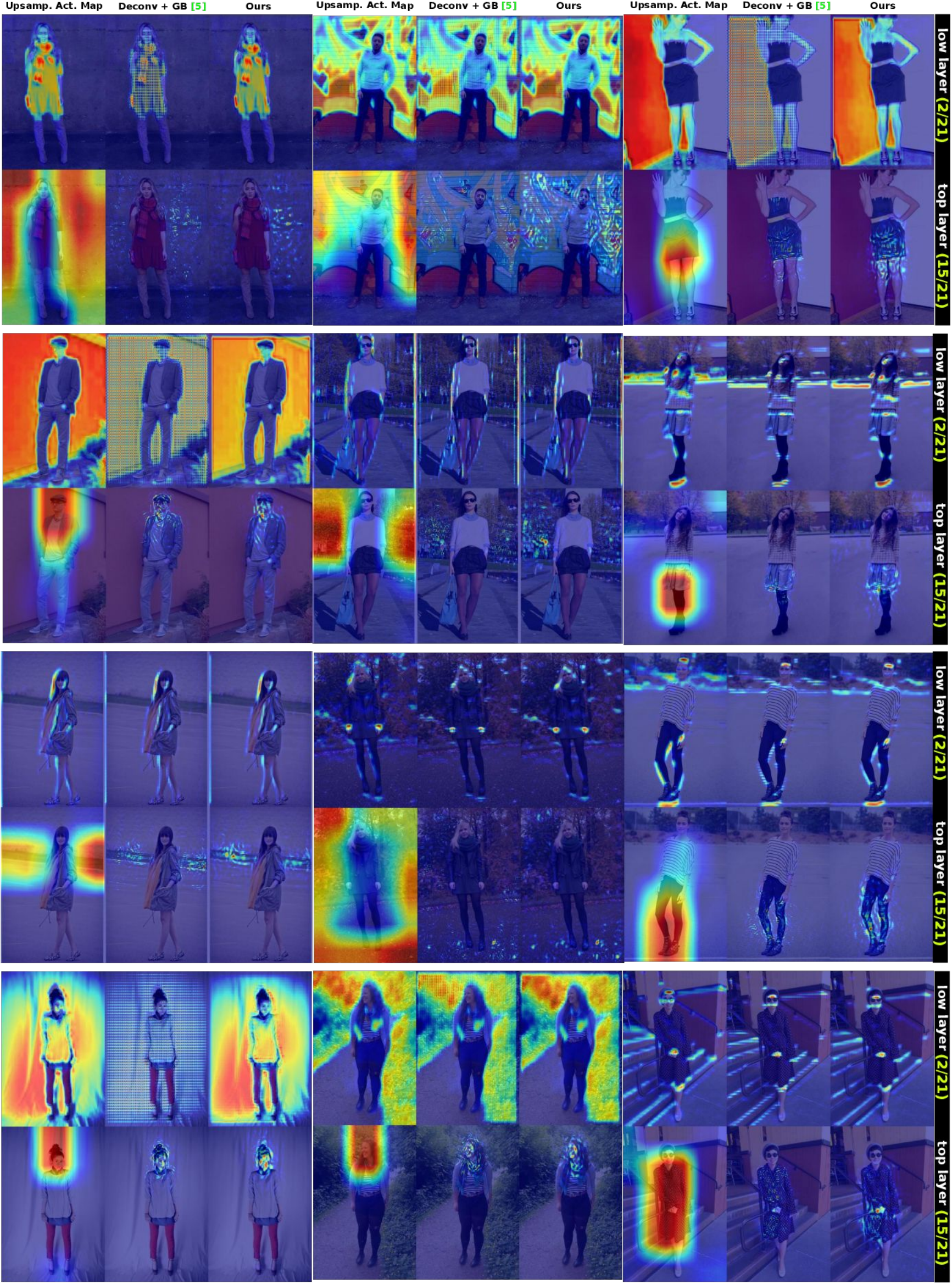}

\caption{
\textbf{Visual quality comparison} for visualizations generated from the \textbf{Fashion114k} dataset~\cite{SimoSerra15}. Note how our heatmaps attenuate the grid-like artifacts introduced by deconvnet-based methods at lower layers. Likewise, our method is able to produce a more detailed visual feedback than upsampled activation maps.
}
\label{fig:visualQualityFashion144k2}
\end{figure*}

\begin{figure*}
\centering
\includegraphics[width=1\textwidth]{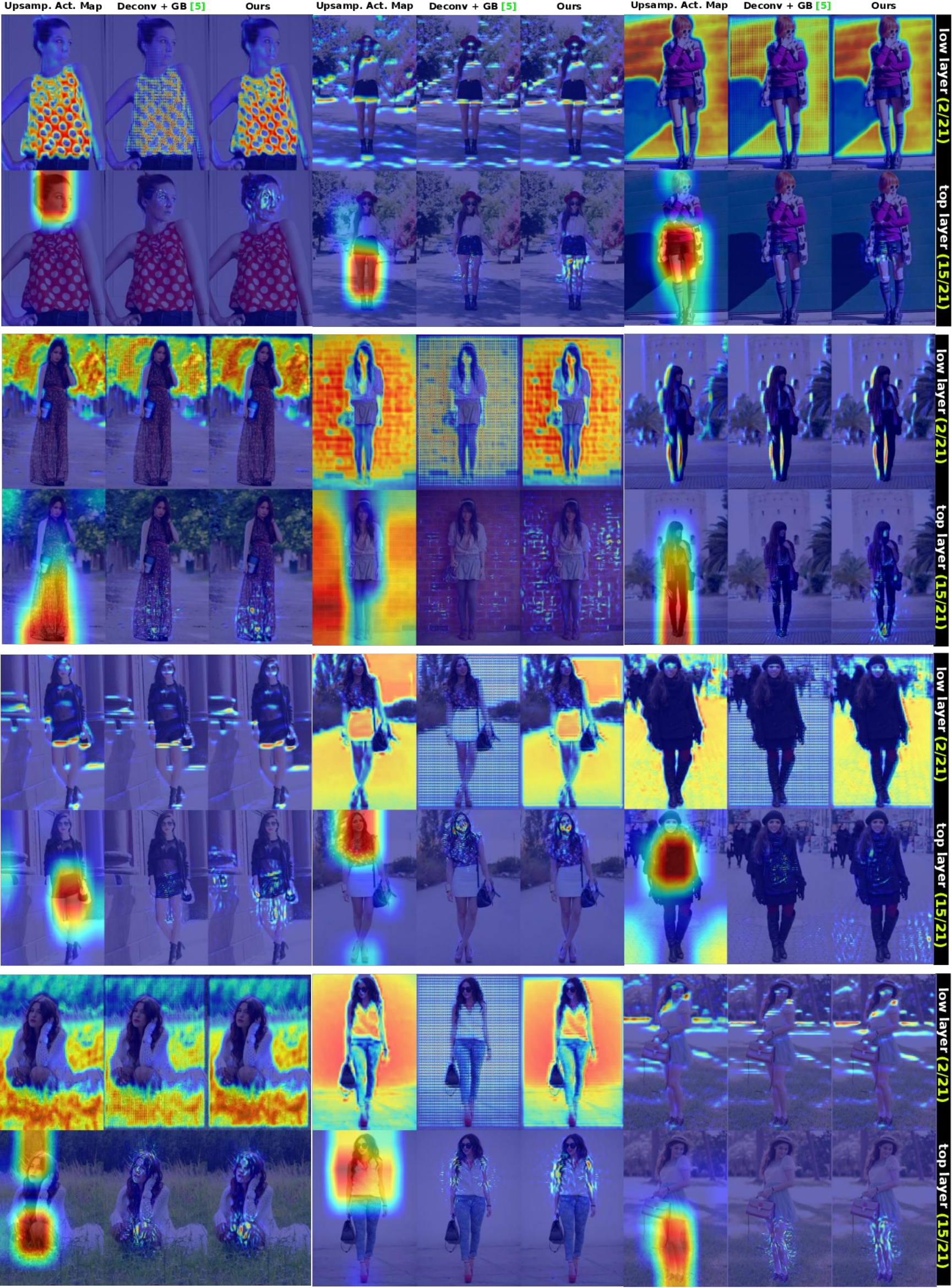}

\caption{
\textbf{Visual quality comparison} for visualizations generated from the \textbf{Fashion114k} dataset~\cite{SimoSerra15}. Note how our heatmaps attenuate the grid-like artifacts introduced by deconvnet-based methods at lower layers. Likewise, our method is able to produce a more detailed visual feedback than upsampled activation maps.
}
\label{fig:visualQualityFashion144k3}
\end{figure*}

We conclude this document by providing extended
results related to the visual quality comparison 
\textit{(Section~4.2 in the original manuscript)}
of the visualizations generated by our method.
Towards this goal, in Figures~\ref{fig:visualQualityImageNetCats1}-\ref{fig:visualQualityFashion144k3} we compare our visualizations 
with upsampled activation maps from internal layers~\cite{netdissect2017,Zhou15Parts} and the 
output of deconvnet combined with guided-backpropagation~\cite{SpringenbergGuidedBack15}.
Following the same methodology as in Figure~2 of the
original manuscript, we focus on visualizations at 
lower and higher layers of the network, i.e., 
layer-2/21 and layer-15/21, respectively.

\begin{figure*}
\centering
\includegraphics[width=1\textwidth]{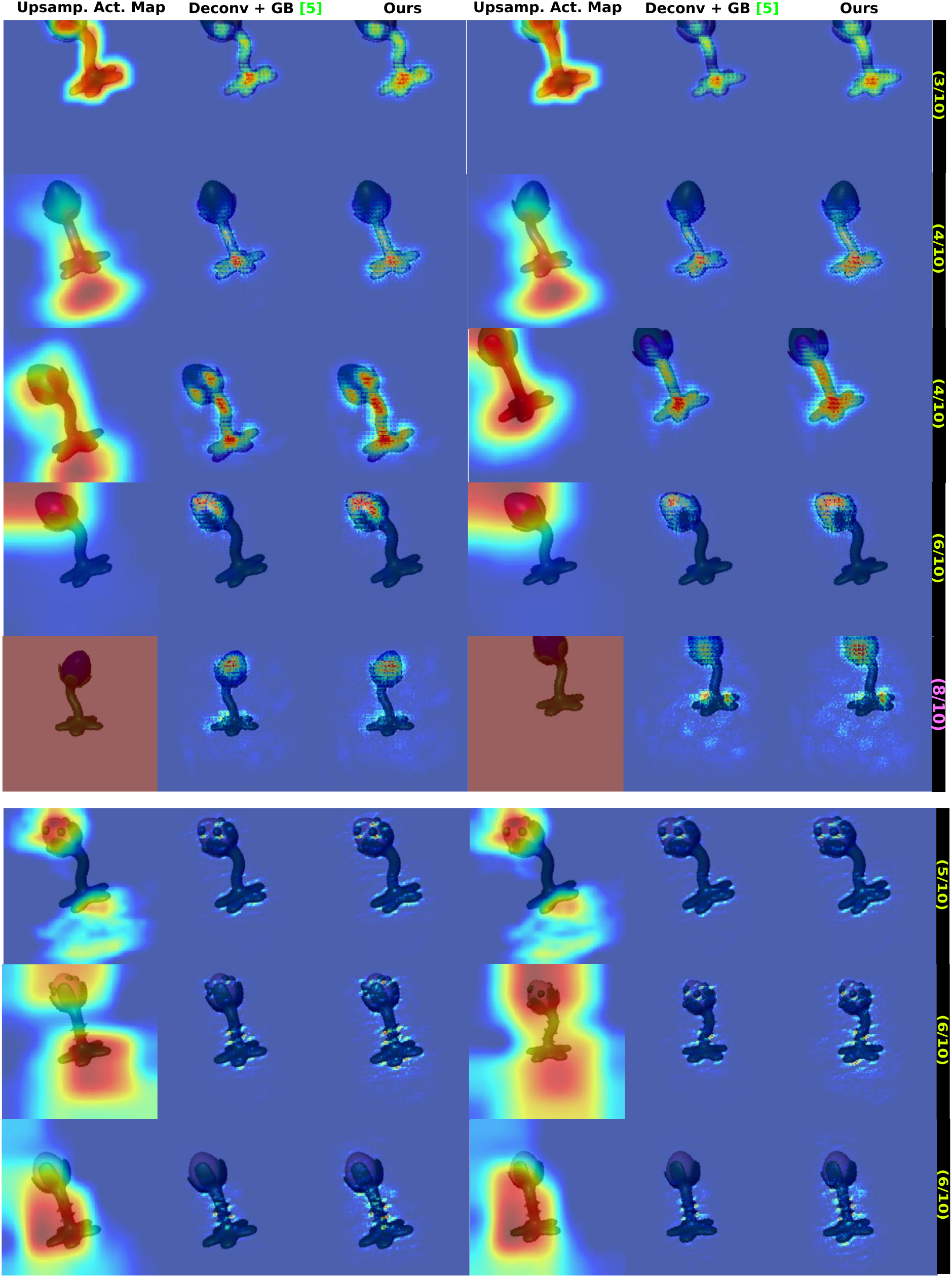}

\caption{
\textbf{Visual quality comparison} for visualizations generated from the \textbf{an8Flower} dataset. The ground truth mask of the first three rows is stem, middle two rows are flowers while the last three rows are balls~(on flower), thorns~(on stem) and both. Note how our heatmaps have a higher coverage and stronger response in the ground truth mask area. Likewise, our method is able to produce a more detailed visual feedback than upsampled activation maps.
}
\label{fig:visualQualityAD}
\end{figure*}

For reference, in Figure~\ref{fig:visualQualityAD} we show similar 
comparisons from explanations generated from models trained 
on the proposed \textit{an8Flower} dataset.

In Figures~\ref{fig:visualQualityImageNetCats1}-\ref{fig:visualQualityFashion144k3} we can corroborate 
that our method to attenuate the grid-like artifacts
introduced by deconvnet methods indeed produces noticeable 
improvements, for lower layers. 
Likewise, for the case of higher layers, the proposed 
method provides a more precise visualization when compared 
to upsampled activation maps.


\newpage

\end{document}